\documentclass[reprint, aps, pre, floatfix]{revtex4-2}

\usepackage{amsmath}
\usepackage{amssymb}
\usepackage{amsthm}
\usepackage{graphicx}
\usepackage{placeins}
\usepackage{nicefrac}
\usepackage{xcolor}
\usepackage{romannum}
\usepackage{soul}
\usepackage{braket}
\usepackage{url}
\usepackage{amsmath,amsfonts,amssymb,amsthm}

\newcommand{\E}{\mathbb{E}}
\newcommand{\B}{\mathcal{B}}
\newcommand{\Var}{\mathrm{Var}}
\newcommand{\Cov}{\mathrm{Cov}}
\newcommand{\tr}{\mathrm{tr}}
\newcommand{\R}{\mathbb{R}}

\newcommand{\op}{\mathrm{op}}
\newcommand{\HS}{\mathrm{HS}}
\newcommand{\defeq}{:=}
\newcommand{\T}{\top}
\newcommand{\Cbar}{\overline{C}}
\newcommand{\bigO}{\mathcal{O}}

\newcommand{\CbarBm}{\overline{\mathbf C}^{\mathcal B}}     
\newcommand{\cbarBm}{\overline{C}^{\mathcal B}}              

\theoremstyle{plain}
\newtheorem{theorem}{Theorem}[section]
\newtheorem{lemma}[theorem]{Lemma}

\newtheorem{assumption}{Assumption}
\newtheorem{property}{Property}

\theoremstyle{remark}

\AtBeginDocument{\pagenumbering{arabic}}

\makeatletter
\DeclareFontFamily{OMX}{MnSymbolE}{}
\DeclareSymbolFont{MnLargeSymbols}{OMX}{MnSymbolE}{m}{n}
\SetSymbolFont{MnLargeSymbols}{bold}{OMX}{MnSymbolE}{b}{n}

\DeclareFontShape{OMX}{MnSymbolE}{m}{n}{
    <-6>  MnSymbolE5
   <6-7>  MnSymbolE6
   <7-8>  MnSymbolE7
   <8-9>  MnSymbolE8
   <9-10> MnSymbolE9
  <10-12> MnSymbolE10
  <12->   MnSymbolE12
}{}

\DeclareFontShape{OMX}{MnSymbolE}{b}{n}{
    <-6>  MnSymbolE-Bold5
   <6-7>  MnSymbolE-Bold6
   <7-8>  MnSymbolE-Bold7
   <8-9>  MnSymbolE-Bold8
   <9-10> MnSymbolE-Bold9
  <10-12> MnSymbolE-Bold10
  <12->   MnSymbolE-Bold12
}{}

\let\llangle\@undefined
\let\rrangle\@undefined
\DeclareMathDelimiter{\llangle}{\mathopen}{MnLargeSymbols}{'164}{MnLargeSymbols}{'164}
\DeclareMathDelimiter{\rrangle}{\mathclose}{MnLargeSymbols}{'171}{MnLargeSymbols}{'171}
\makeatother

\usepackage{silence}
\definecolor{darkblue}{rgb}{0,0,0.6}
\definecolor{darkred}{rgb}{0.6,0,0}
\definecolor{darkgreen}{rgb}{0,0.6,0}

\usepackage[colorlinks=true, 
            urlcolor=darkblue, 
            citecolor=darkblue, 
            linkcolor=darkred, 
            hyperfootnotes=false]{hyperref}

\begin{document}

\author{Jaeyong Bae}
\affiliation{Department of Physics, Korea Advanced Institute of Science and Technology, Daejeon 34141, Republic of Korea}

\author{Hawoong Jeong}
\email{hjeong@kaist.edu}
\affiliation{Department of Physics, Korea Advanced Institute of Science and Technology, Daejeon 34141, Republic of Korea}
\affiliation{Center of Complex Systems, Korea Advanced Institute of Science and Technology, Daejeon 34141, Republic of Korea}

\title{Gaussian Universality in Neural Network Dynamics with Generalized Structured Input Distributions}

\begin{abstract}
Analyzing neural network dynamics via stochastic gradient descent (SGD) is crucial to building theoretical foundations for deep learning. Previous work has analyzed structured inputs within the \textit{hidden manifold model}, often under the simplifying assumption of a Gaussian distribution. We extend this framework by modeling inputs as Gaussian mixtures to better represent complex, real-world data. Through empirical and theoretical investigation, we demonstrate that with proper standardization, the learning dynamics converges to the behavior seen in the simple Gaussian case. This finding exhibits a form of universality, where diverse structured distributions yield results consistent with Gaussian assumptions, thereby strengthening the theoretical understanding of deep learning models.
\end{abstract}
\maketitle

\section{Introduction}

The study of artificial neural networks through the lens of statistical physics has a well-established history. Neural networks trained on samples from a distribution have traditionally been analyzed as optimization problems within complex systems \citep{review}. These problems are generally classified based on the learning process, whether the entire dataset or a subset is used per iteration, as in gradient descent and batch gradient descent, respectively, or whether a single sample is used per iteration, as in stochastic gradient descent (SGD, also known as on-line learning).\\
When the entire dataset is used, one can fix the dataset size and interpret the neural network and corresponding loss function as analogous to a spin system and potential energy. The replica method can then be employed to analyze the properties of optimally trained networks and their capabilities, such as learnability. This approach has provided successful interpretations of simple one- or two-layer perceptrons during the early development of neural networks \citep{replica1, replica2, replica3, replica4}, and more recently has demonstrated applicability to more advanced settings including generative models \citep{gentheo, attentiontheo}.\\
On the other hand, when considering SGD, where a single dataset is used per iteration, the neural network variables are updated to optimize the loss for each sample. The behavior of neural networks under such independent random sampling can be described without the need for the replica method. By analyzing the equations of motion derived from this approach, it is possible to track generalization error and the time evolution of low-dimensional order parameters describing the evolution of the weights in the neural network \citep{on1, on2, on3, on4, on5, on6, on7}.\\
Recently, studies have shifted focus toward understanding neural network behavior under structured input data. The notion of structured data posits that despite the high-dimensional nature of typical datasets (such as MNIST in \citet{deng2012mnist} with 28$\times$28 dimensions and CIFAR in \citet{CIFAR10} with 3$\times$32$\times$32 dimensions), they can often be distilled into lower-dimensional representations. For instance, in the MNIST dataset, the digits display structured patterns like lines and curves rather than random pixel arrangements. This characteristic of low-dimensional structural features in data has been discussed in numerous studies \citep{intrin1, intrin2, intrin3, intrin4, gan1, gan2}.\\

Such structured inputs can be modeled in various ways. One prominent framework is the \textit{hidden manifold model}, which assumes that high-dimensional inputs arise from low-dimensional latent variables passed through nonlinear transformations \citep{goldt2020modeling, goldt2022gaussian}. Within this framework, it has been shown that the preactivation distributions of neurons approach Gaussian in certain limits, a phenomenon often referred to as the \textit{Gaussian equivalence property}. This property enables tractable analysis of deep learning dynamics.

Other studies incorporate input heterogeneity, such as sparse or spiked structures, using random matrix theory to analyze neural network behavior \citep{spike1,spike2,spike3}. 
In these cases, various forms of conditional Gaussian equivalence have also been identified, providing theoretical insights~\citep{spikelear1,spikelear2,spikelear3,spikelear4}.

Despite such progress, existing studies on the hidden manifold and spiked models remain constrained to specific settings. Recent work has aimed to generalize these models, developing solvable frameworks that further establish universality laws for structured data, and extending these efforts to understand the role of non-Gaussian inputs in analytically tractable scenarios \citep{univer1,univer2,univer3,univer4,univer5,univer6, nonG1, nonG2, nonG3}. Nevertheless, these approaches remain limited in capturing the full complexity of real-world data, which are often better represented by general Gaussian mixtures \citep{whygm1, whygm2, whygm3}.

Motivated by this, we focus on extending the hidden manifold model to the case of Gaussian mixture inputs. Our goal is to examine whether neural network dynamics under such generalized input distributions still converge to those observed in the simple Gaussian case, thus reinforcing the broader applicability of Gaussian universality.

We perform numerical simulations across diverse input distributions, including general Gaussian mixtures under mild restrictions. Varying the input distribution, we chart the regimes where SGD training trajectories coincide with their Gaussian counterparts. We find consistent evidence that, after standardization and under sufficiently weak correlations, preactivation statistics approach their Gaussian counterparts and the resulting dynamics align with the Gaussian case. When the dynamics deviate, we observe correlations with shifts in input level statistics or with distributional distances to a Gaussian baseline, suggesting that these quantities may serve as broad diagnostics.

Our key findings are as follows:
\begin{itemize}
\item Standardization of input datasets modeled by Gaussian mixtures in low dimensions results in convergence with the dynamics observed under Gaussian inputs.
\item The observed convergence is largely due to the weakly correlation condition of dataset and model settings. With the standardization, weakly correlated random variables combinations converge to Gaussian, even under general distribution.
\item In the context of the \textit{Gaussian equivalence property}, our analysis suggests that, even under Gaussian-mixture inputs, the preactivation distribution remains boundedly close to its Gaussian counterpart; the tightness of this bound is affected by the effective degree of weak correlation.

\end{itemize}
This study is organized as follows. In Section~\ref{Background} we provide a brief background on relevant studies, and in Section~\ref{Method} we describe the Gaussian mixture settings and experimental conditions. In Section~\ref{Results}, we present the experimental results showcasing the convergence patterns. Then in Section~\ref{Derivation} we offer a mathematical proof of the observed phenomena. In Section~\ref{Discussion} we discuss limitations and possible expansion of our results. Finally, we conclude our study in Section~\ref{Conclusion}.

\begin{figure*}[t]
\centerline{\includegraphics[width=0.8\textwidth]{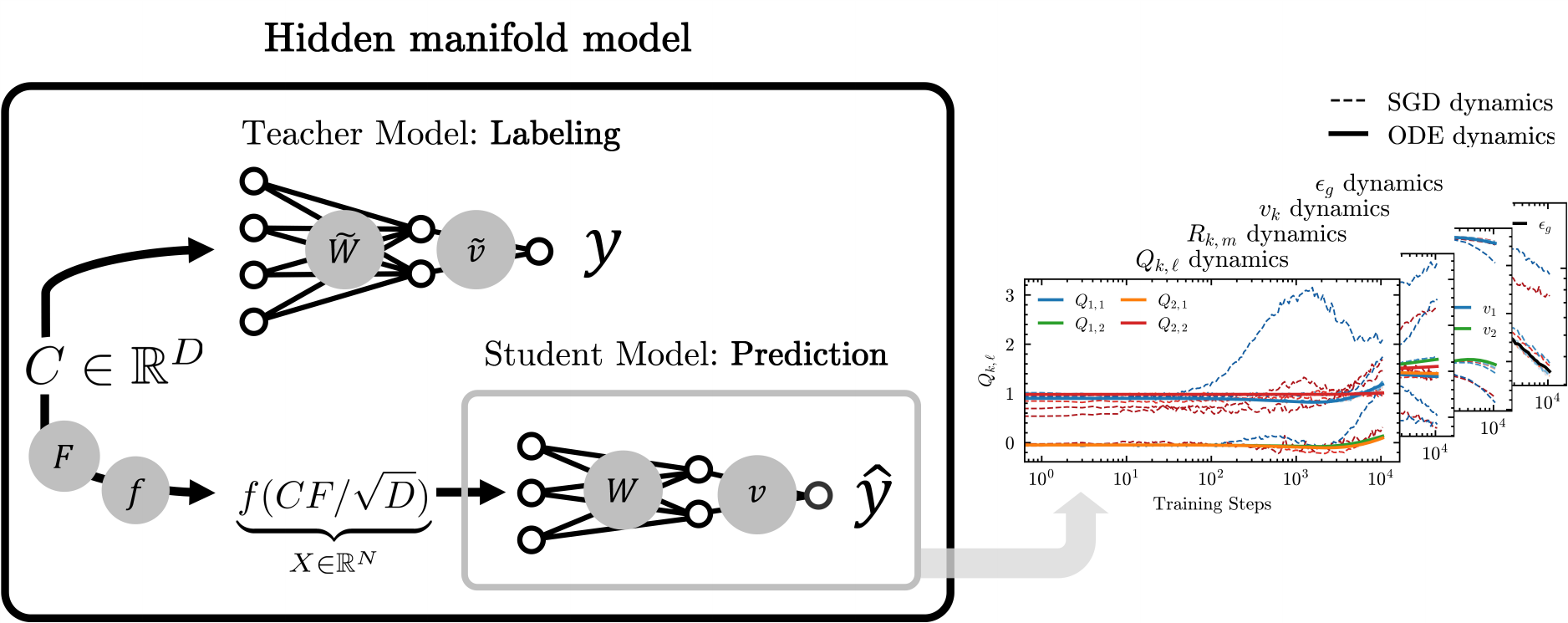}}
\caption{Illustration of the hidden manifold model (\citet{goldt2020modeling}) and our experimental scheme. The term \textit{ODE dynamics} refers to the outcomes from ordinary differential equation (ODE) simulations, which align with SGD under a simple Gaussian input, since ODE represents the continuous-time limit of SGD under Gaussian input. In contrast, \textit{SGD dynamics} describes the results obtained from running SGD with $C$ from various distributions. We consider $C$ drawn from a simple Gaussian, from heavy-tailed distributions such as the Lorentz distribution, and from general Gaussian mixtures as described in Sec.~\ref{Method}}
\label{figure1}
\end{figure*}

\section{Background}\label{Background}
This Section~\ref{Background} offers an overview of previous findings from \citet{goldt2020modeling}. From the teacher-student model framework and one of its evolved variants, the hidden manifold model. We introduce their findings in neural network dynamics.

\subsection{Hidden Manifold Model}\label{Background:HiddenManifold}
The teacher-student model is a well-regarded method in the study of high-dimensional problems \citep{gabrie2020mean, baity2018comparing, bahri2020statistical, zdeborova2020understanding, goldt2019dynamics}. This model framework consists of a teacher model that generates dataset labels and a student model that learns the labels. In a two-layer neural network, the weights of the first and second layers of the teacher model are represented by matrices $\widetilde{W} \in \mathbb{R}^{M\times D}$ and $\widetilde{v} \in \mathbb{R}^{1\times M}$, respectively. We define the activation function of the teacher model as $\widetilde{g}$. Similarly, the weights of the first and second layers of the student model are denoted by $W \in \mathbb{R}^{K\times N}$ and $v \in \mathbb{R}^{1 \times K}$, with the activation function represented as $g$.\\
In the canonical teacher-student model, the input $X \in \R^N$ is typically element-wise i.i.d. from a Gaussian distribution. 
But here, we aim for input characteristics that reflect intrinsic structural properties rather than being merely extrinsically Gaussian.\\
To embed these intrinsic properties into the input $X$, \citet{goldt2020modeling} utilized a $D$-dimensional vector, $C \in \R^D$, that follows an element-wise i.i.d. Gaussian distribution. This is achieved through the feature matrix $F \in \R^{D \times N}$ and nonlinear function $f$ as follows:
\begin{equation}
    X = f(CF/\sqrt{D}) = f(U)\in \R^{N}.
\end{equation}
By modeling the dataset in this manner, the input $X$ intrinsically reflects the characteristics of $C$, which is distributed as Gaussian. Furthermore, the labels generated by the teacher model are derived not directly from $X$ but from $C$, which reflects the dominance of the intrinsic characteristics in the true label.\\
In summary, the teacher-student model results can be expressed as follows:
\begin{equation}\label{teacherstudent}
    y = \widetilde{g}(C\widetilde{W}^\top /\sqrt{D})\widetilde{v}^\top, \quad \hat{y} = {g}(X{W}^\top /\sqrt{N}){v}^\top.
\end{equation}
This model, recognizing a hidden structure in lower dimensions, is termed the hidden manifold model \citep{goldt2020modeling}; Fig.~\ref{figure1} shows a schematic of the model along with our experimental scheme (see Section~\ref{Results}).
For the convenience of subsequent discussions, we denote the preactivations of the teacher and student models as $\nu$ and $\lambda$, respectively:
\begin{align}\label{nulambda}
     \nu &= C\widetilde{W}^\top/\sqrt{D} \in \R^M \nonumber,\\ 
    \lambda &= XW^\top/\sqrt{N} = f(U)W^\top/\sqrt{N} \in \R^K.  
\end{align}
When input $C$ spans beyond a singular data point to represent a dataset of size $P$, it assumes a matrix form, denoted as $C \in \R^{P \times D}$. Consequently, each preactivation is expressed through matrices $\nu \in \R^{P \times M}$ and $\lambda \in \R^{P \times K}$. \\
Without any additional notice, the notation $\mathsf{M}_{i,j}$ represents the element located in the $i$-th row and $j$-th column of any given matrix $\mathsf{M}$, and $\mathsf{v}_i$ denotes the $i$-th component of any vector $\mathsf{v}$.\\
Below, we frequently consider an infinitely large dataset dimension $N$ and intrinsic dimension $D$ ($N \to \infty, D \to \infty$), a scenario often referred to as the thermodynamic limit from the perspective of statistical physics. To maintain consistency with prior studies, this research also adopts the term thermodynamic limit to describe the $N \to \infty$, $D \to \infty$, when $D/N$ is finite scenario.

\subsection{Dynamics of Neural Network Weights}\label{sgd}

In this study, we adopt the approach of \citet{goldt2020modeling} to update the weights of the student model using a scaled SGD under quadratic loss $\mathcal{L}=1/2 (\hat{y}-y)^2$, with a learning rate $\eta$. The updates are performed while considering the vector form of $\lambda$, $\nu$ and $U$,
\begin{equation}\label{eqn:sgd}
    \begin{aligned}
        W_{k,i} & :=W_{k,i}-\frac{\eta}{\sqrt{N}} v_k (\hat{y}-y) g^{\prime}\left(\lambda_k\right) f(U_i), \\v_k &:= v_k -\frac{\eta}{N} g\left(\lambda_k\right) (\hat{y}-y).
    \end{aligned}
\end{equation}
By defining the normalized number of steps as $t = 1/N$ in the thermodynamic limit $N \to \infty$, which can be interpreted as a continuous time-like variable, the updates transform into ordinary differential equations (ODEs).\\
For example, $v_k$ satisfies the following ODE:
\begin{equation}\label{eqn:v}
    \frac{dv_k}{dt} = \eta\left[ \sum^M_m \widetilde v_m I_2(k,m) -\sum^K_j  v_j I_2(k,j) \right ],
\end{equation}
where $I_2(k,m) = \mathbb{E}[g(\lambda_k) \widetilde{g}(\nu_m)]$ and $I_2(k,j) = \mathbb{E}[g(\lambda_k){g}(\lambda_j) ]$ represent the correlations of functions.\\
Using a similar approach for $v$, we can derive the dynamics of our teacher-student model in the form of ODEs, as detailed in Appendix~\ref{section:background}. The dynamics are predominantly influenced by the correlations of specific functions, such as $I_2(k,m)$,  and $I_2(k,j)$. To calculate these correlations of functions, referred to here as the \textit{function correlation}, we require information on the underlying distribution of $\{\lambda, \nu\}$.
\subsection{The Gaussian Equivalence Property}
\citet{goldt2020modeling} investigated the distribution of preactivations in hidden manifold model, $\{\lambda, \nu\}$ under certain assumptions. To summarize the previous findings, $\{\lambda, \nu\}$ follow a Gaussian distribution characterized by a specific covariance matrix.\\
Assuming that the weights $W, \widetilde{W}$ and feature matrix $F$ do not significantly alter the input properties, having elements in $\mathcal{O}(1)$ scale, adopt the following assumptions
\begin{assumption}\label{assumption:goldt:F}
The feature matrix has bounded property,
\begin{equation}
        \frac{1}{\sqrt{D}}\sum_{r=1}^D F_{r,i} F_{r,j} = \mathcal{O}(1), \quad \sum_{r=1}^D F_{r,i}^2 = D,
\end{equation}
\end{assumption}
and
\begin{assumption}\label{assumption:goldt:bounded}
For all $p,q \geq 1$ and any indices $k_1, \cdots, k_p, r_1, \cdots, r_q$:
    \begin{equation}
        \cfrac{1}{\sqrt{N}}\sum_i W_{k_1,i} \cdots  W_{ k_p,i} \times F_{r_1,i} \cdots F_{r_q,i} = \mathcal{O}(1),
    \end{equation}
with the $q$ and $p$ distinct. A similar scaling holds for terms involving the teacher weights.
\end{assumption}
Under theses conditions taken from \citet{goldt2020modeling}, we can approximately calculate the covariances of $\{\lambda, \nu\}$, i.e., $\mathbb{E}\left[\lambda_k \lambda_\ell\right], \mathbb{E}[\lambda_k\nu_m], \mathbb{E}[\nu_m \nu_n]$.
This analysis enables the derivation of the asymptotic form of all covariance matrices of preactivations $\{\lambda, \nu\}$, where higher-order correlations vanish in the thermodynamic limit. The hidden manifold model setup (Section~\ref{Background:HiddenManifold}), e.g., the teacher model, the student model, or the feature matrix, satisfies these assumptions as shown by \citet{goldt2020modeling}.\\
Consequently, the preactivations follow a Gaussian distribution, a result summarized as the Gaussian equivalence property (GEP).
\begin{property}[Gaussian equivalence property (GEP)]\label{GEP}
In the thermodynamic limit ($N\to \infty$, $D \to \infty$), with finite $K$, $M$, $D/N$, and under Assumption~\ref{assumption:goldt:F} and Assumption~\ref{assumption:goldt:bounded}, if $C$ follows a standard Gaussian distribution $\mathcal{N}(0,I)$, then $\{\lambda, \nu \}$ are jointly Gaussian variables of dimension $K+M$. This means that the statistics involving $\{\lambda, \nu\}$ are entirely represented by their mean and covariance.
\end{property}
Property~\ref{GEP} taken from \citet{goldt2020modeling} states that the preactivation distribution $\{\lambda, \nu \}$ follows a joint Gaussian distribution.
This property allows us to understand the student model's dynamics (e.g., student weight, generalization error) via the mean and covariance of the joint Gaussian distribution of preactivations. For example, the dynamics of the student weight (Eq.~(\ref{eqn:v})) is tractable by calculating $I_2$ under the $\{\lambda, \nu \}$ distribution. Detailed derivations are provided in Appendix~\ref{section:background}.\\
Before expressing the covariance matrix, for convenience we redefine $\overline{\lambda}_k$ as
\begin{equation}
    \overline{\lambda}_k = \cfrac{1}{\sqrt{N}} \sum_{i=1}^N W_{k,i} ( f(U_i)-\mathbb{E}_{u \sim \mathcal{N}(0,1)}[f(u)]).
\end{equation}
$\overline{\lambda}_k$ also follows a jointly Gaussian distribution with $\mathbb{E}[\overline{\lambda}_k]=0$.
Consequently, the new distribution $\{\overline{\lambda}, \nu\}$ has means of \begin{equation}
\mathbb{E}\left[\overline{\lambda}_k\right]=\mathbb{E}\left[\nu_m\right]=0,
\end{equation}\label{mean:preactivation}
and covariances as follows:
\begin{align}
Q_{k, \ell} &\equiv \mathbb{E}\left[{\overline\lambda}_k {\overline\lambda}_{\ell}\right]=\left(c-a^2-b^2\right) \Omega_{k, \ell}+b^2 \Sigma_{k, \ell}, \label{eqn:Q}\\R_{k, m} &\equiv \mathbb{E}\left[\overline{\lambda}_k \nu_m\right]=b \frac{1}{D} \sum_{r=1}^D S_{k,r} \widetilde{W}_{m,r}, \label{eqn:R}\\T_{m, n}& \equiv \mathbb{E}\left[\nu_m \nu_n\right]=\frac{1}{D} \sum_{r=1}^D \widetilde{W}_{m,r} \widetilde{W}_{n,r}\label{eqn:T}.
\end{align}
Here, $a$, $b$, and $c$ represent the statistical properties of the nonlinear function $f$, used in the transformation of student model inputs $X$, $X=f(CF/\sqrt{D})$, given as
\begin{equation}
    a = \mathbb{E}[f(u)], \quad b = \mathbb{E}[uf(u)], \quad c = \mathbb{E}[f(u)^2],
\end{equation}
under ${u \sim \mathcal{N}(0,1)}$.
The newly defined matrices satisfy the following relations:
\begin{align}
S_{k,r} & \equiv \frac{1}{\sqrt{N}} \sum_{i=1}^N W_{k,i} F_{r,i}, \\\Omega_{k, \ell} & \equiv \frac{1}{N} \sum_{i=1}^N W_{k,i} W_{\ell,i}, \\\Sigma_{k, \ell} & \equiv \frac{1}{D} \sum_{r=1}^D S_{k,r} S_{\ell, r}.
\end{align}
For compact notation, we focus on the symmetric nonlinear function $f$ that satisfies $a = \mathbb{E}[f(u)]=0$. These defined covariances capture the essential characteristics of the teacher-student model dynamics.\\
To summarize, within the context of the hidden manifold model characterized by a simple Gaussian distribution, the learning dynamics of the student model are primarily influenced by the function correlation terms. Such terms, which depend on the distributional properties of $\{\lambda, \nu\}$, can be calculated once the distribution is determined. Under Assumption~\ref{assumption:goldt:F} and Assumption~\ref{assumption:goldt:bounded}, the GEP shows that the preactivations $\{\lambda, \nu\}$ follow a Gaussian distribution, allowing us to analytically dissect the dynamics of the student model.

\section{Method}\label{Method}

In this section, we detail our approach to configuring inputs and our experimental settings, where we consider a two-layer teacher-student model as in Section~\ref{section:background}.

\subsection{Dimension-wise Gaussian Mixture Setting}

\begin{figure}[htb]
    \centering
    \includegraphics[width=\linewidth]{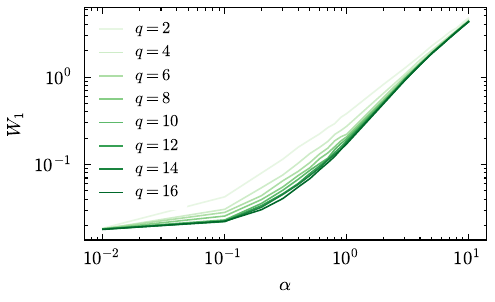}
    \caption{Wasserstein--1 distance $W_{1}(\mathcal{P},\mathcal{N})$ for different values of $\alpha$ and number of components $q$, where $\mathcal{P}$ represents the dimension-wise Gaussian mixture distribution.}
    \label{fig:w1_dist}
\end{figure}

In this study, we employ a Gaussian mixture distribution as the teacher model input. To distinguish between Gaussian inputs and Gaussian mixture inputs, we denote Gaussian mixture inputs as ${C^{\mathcal{M}}}$.
For each dimension $r\in{1,\ldots,D}$, let $C^{\mathcal{M}}_r$ be drawn from an independent $q$ component Gaussian mixture. Component means are sampled i.i.d. from the uniform distribution on $[-\alpha,\alpha)$, standard deviations are sampled i.i.d. from $(0,\beta)$, and mixing weights $\pi_k$ are normalized so that $\sum_{k=1}^q \pi_k=1$. The marginal density for dimension $r$ is
\begin{equation}
p_r(x)=\sum_{k=1}^q \pi_k, \mathcal{N}(x;\mu_k,\sigma_k^2), \qquad r=1,2,\ldots,D .
\end{equation}
Then we generate $C^{\mathcal{M}}\in \R^{P \times D}$ by sampling from each respective Gaussian mixture.
This methodology allows us to conduct empirical investigations across a spectrum of Gaussian mixtures by adjusting the parameters $\alpha$, $\beta$ and $q$.
In Fig.~\ref{fig:w1_dist}, we check that increasing $\alpha$ under fixed $\sigma_k=1$ leads to a monotonically increasing Wasserstein--1 distance from a single Gaussian distribution. 

\subsection{Block-dependent Gaussian Mixture Setting}
To deal with internal dependence, we adopt a block-dependent Gaussian mixture model.
Within each block $\mathcal{B}_b$ of size $s_b \defeq |\mathcal{B}_b|$, the data are generated from a $q$-component Gaussian mixture model. First, a component label $\kappa_b \in \{1, \dots, q\}$ is drawn with probability $P(\kappa_b = k) = \pi_k$. Conditional on this label, the vector $C_{\mathcal{B}_b}$ is drawn from a $s_b$ dimension multivariate normal distribution:
\begin{equation}
C_{\mathcal{B}_b} \mid (\kappa_b = k) \sim \mathcal{N}(m_{b,k}, \Sigma_{b,k}),
\label{eq:block_mixture_model}
\end{equation}
where the mean and covariance have a specific structure
\begin{equation}
\begin{aligned}
m_{b,k} &= \mu_{b,k} \mathbf{1} + \delta_{b,k} u_b, \\
\Sigma_{b,k} &= D_b \left( (1 - \rho_{b,k} - \tau_{b,k})I + \rho_{b,k} \mathbf{1}\mathbf{1}^\T + \tau_{b,k} v_b v_b^\T \right) D_b.    
\end{aligned}
\label{eq:block_params}
\end{equation}
Here, $\mathbf{1}$ is the vector of ones of length $s_b$, and $D_b = \mathrm{diag}(\sigma_{b,r})_{r \in \mathcal{B}_b}$ is a diagonal matrix of coordinate-wise standard deviations. 
The term $\delta_{b,k}u_b$ introduces mean perturbation along $u_b$, while the covariance combines an equicorrelated component with strength $\rho_{b,k}$ and a rank-1 perturbation $\tau_{b,k}v_b v_b^\T$.
We then generate $C^{\mathcal{B}} \in \R^{P \times D}$ by sampling from the block-dependent Gaussian mixture.
Consequently, $C^{\mathcal{B}}$ constitutes a block-dependent Gaussian–mixture dataset. Each block $\mathcal{B}_b$ is drawn from its own $q$ component $s_b$ dimensional Gaussian mixture with block specific component weights, mean shift, and covariance and is independent across blocks.

\subsection{Additional Experimental Settings}
For robustness, we use the dimension of student model input $X$ was set to $N=1024$, $2048$, $4096$, and the dimension of teacher model input $\mathcal{C}$ was set to $D=N\delta$, where $\delta=1/2$, $1/4$, $1/8$, $1/16$. The dimensions of the hidden layers for both teacher and student models were uniformly set to $K=M=2$. Both the teacher and student models employed the same activation function, $g(x)=\widetilde{g}(x)=\text{ReLU}(x)$. The nonlinear function $f(x)$ used to generate the student input was $f(\cdot)=\text{tanh}(\cdot)$.\\
Because different $(N,D)$ pairs correspond to distinct teacher-student configurations, it is impractical to present all results within a single view. Accordingly, the main text focuses on the large-system setting $N=4096$ with $\delta=1/2$; results for other parameter choices are provided in the Appendix.
The learning rate was set to $\eta=0.2$, and training was conducted using the quadratic loss $\mathcal{L}=\frac{1}{2} (\hat{y}-y)^2$ with a scaled SGD update procedure as shown in Eq.~(\ref{eqn:sgd}). The student model was updated for a total of $10 \times N$ steps. In same $(N, D)$ pair, we used the same initial conditions ($\widetilde{W}, \widetilde{v}, W, v$ and $F$) to obtain the evolution of the dynamics.

\begin{figure*}[htb]
\centerline{\includegraphics[width=\textwidth]{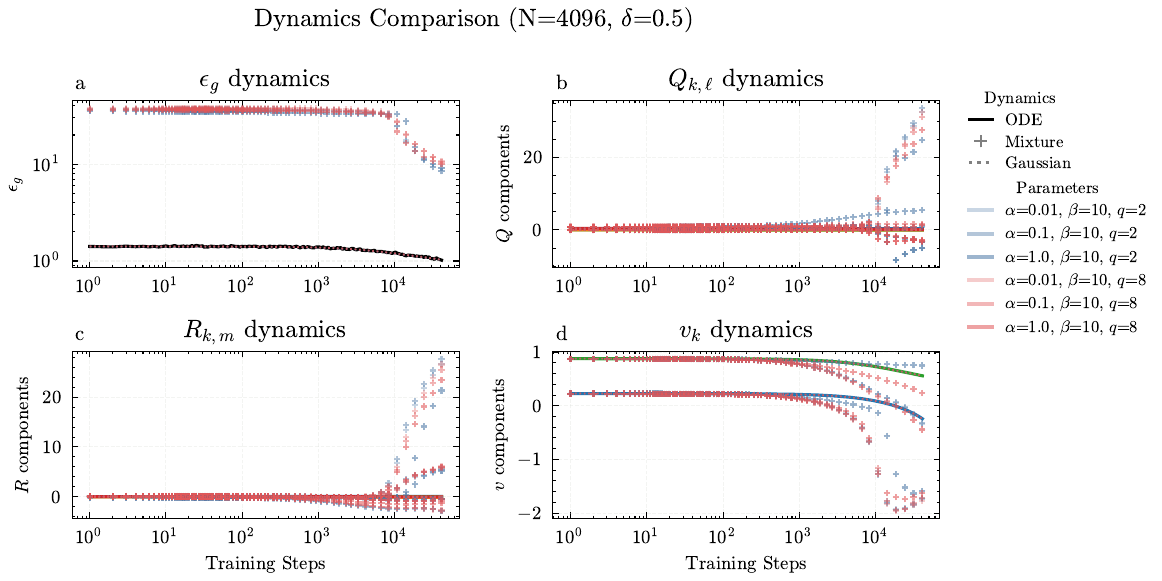}}
\caption{Examples of dynamics under unstandardized Gaussian mixtures with $q=2, q=16$, and $\alpha=0.01, 0.1, 1$, $\beta = 10$. Dynamics of (a) generalization error $\epsilon_g$, (b) covariance matrix $Q$, (c) covariance matrix $R$, and (d) weight of the second layer $v$. The SGD results are averaged over 5 runs. As a baseline, the ODE dynamics (solid line) align perfectly with SGD simulations on simple Gaussian inputs (dotted line). However, a significant divergence occurs between the ODE predictions and the actual SGD dynamics for unstandardized Gaussian mixtures (crosses), demonstrating the failure of the theory in this setting.}
\label{figure3_1}
\end{figure*}

\begin{figure*}[htb]
\centerline{\includegraphics[width=\textwidth]{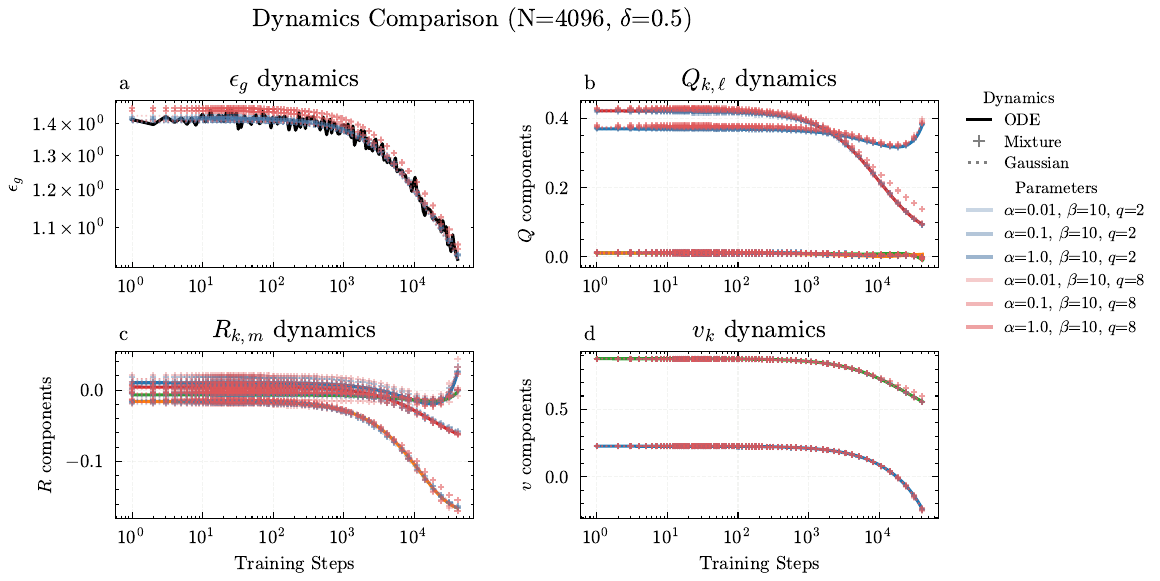}}
\caption{Examples of dynamics under standardized Gaussian mixtures with $q=2, q=16$, and $\alpha=0.01, 0.1, 1$, $\beta = 10$. Dynamics of (a) generalization error $\epsilon_g$, (b) covariance matrix $Q$, (c) covariance matrix $R$, and (d) weight of the second layer $v$. The SGD results are averaged over 5 runs. The close agreement between the ODE predictions (solid line) and SGD simulations (crosses) demonstrates that standardization effectively restores the universal dynamics. This supports the empirical claim that aligning low-order moments is sufficient to recover Gaussian learning dynamics in these settings.}
\label{fig:std_dynamics}
\end{figure*}

\begin{figure*}
    \centering
    \includegraphics[width=\textwidth]{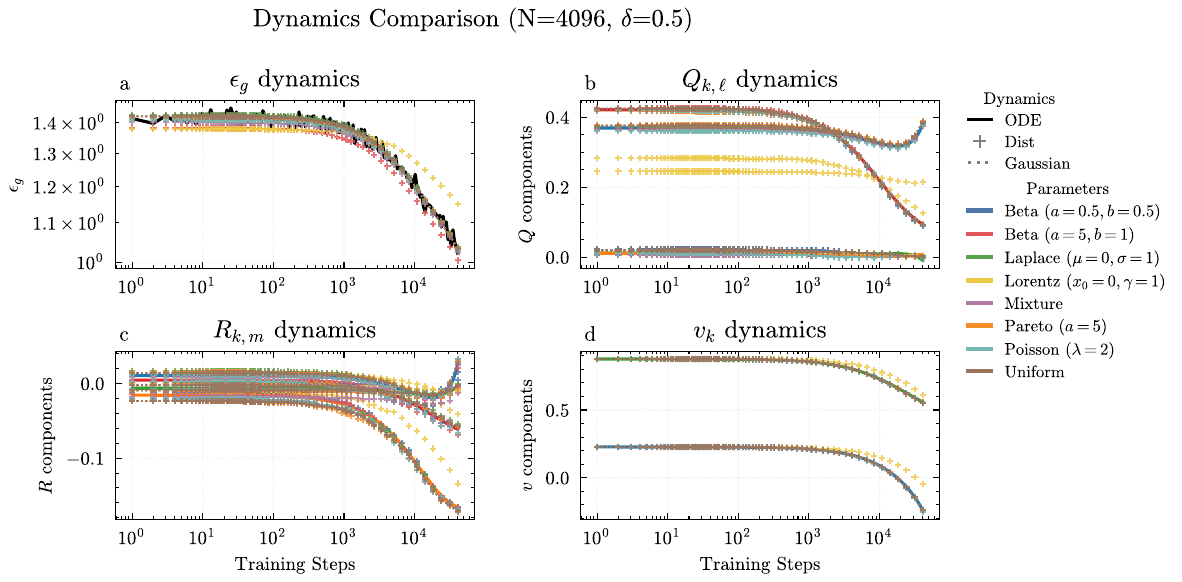}
    \caption{Examples of dynamics under various distribution settings (parameterized in Table~\ref{tab:dist}) after standardization. Dynamics of (a) generalization error $\epsilon_g$, (b) covariance matrix $Q$, (c) covariance matrix $R$, and (d) weight of the second layer $v$. The SGD results are averaged over 5 runs. The Lorentz (Cauchy) distribution is the sole, crucial exception. Its failure to converge, due to its lack of finite moments, provides strong evidence that the universality is specifically governed by the alignment of the moments.}
    \label{fig:dynamics_dist}
\end{figure*}

\begin{figure*}[htb]
\centerline{\includegraphics[width=\textwidth]{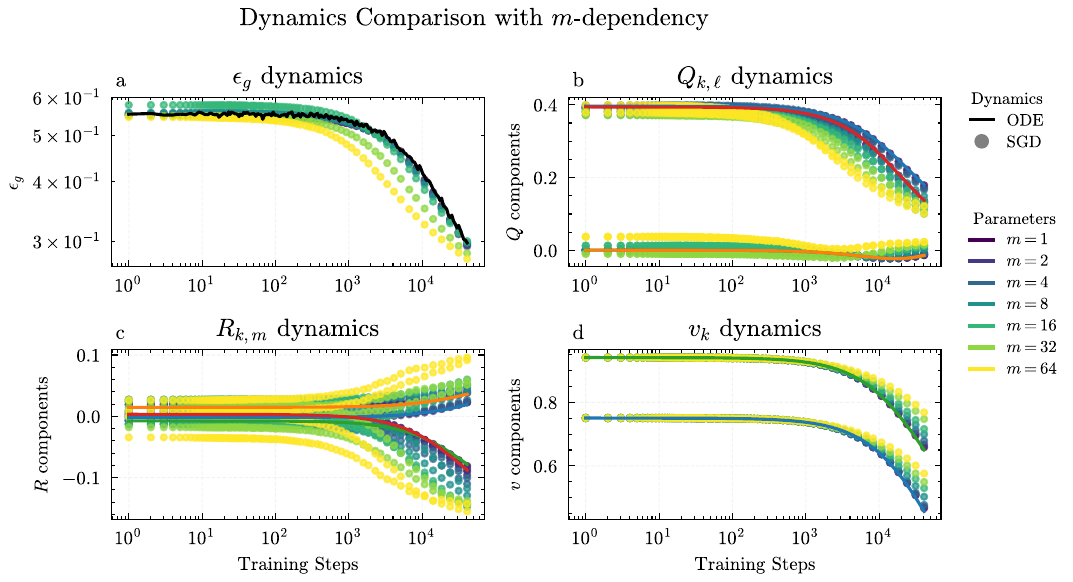}}
\caption{Examples of dynamics under $m$ block dependent Gaussian mixtures with $m=1,2,\ldots,64$, $N=D=4096$. Dynamics of (a) generalization error $\epsilon_g$, (b) covariance matrix $Q$, (c) covariance matrix $R$, and (d) weight of the second layer $v$. The SGD results are averaged over 5 runs. Increasing $m$ induces stronger random correlations and progressively weakens agreement between ODE and SGD. For $m\ll D$ the standardized inputs behave effectively Gaussian and the dynamics coincide; departures grow monotonically with $m$, suggesting an $m/D$-type control of convergence.}
\label{fig:figure_m}
\end{figure*}

\section{Results}\label{Results}

The theoretical framework outlined in Section~\ref{Background} provides a powerful tool for analyzing learning dynamics, but it rests on a critical assumption: a simple Gaussian input distribution. This raises a fundamental question with significant practical implications: \textit{Can this powerful theoretical machinery be applied to more realistic, non-Gaussian data?} If so, under what conditions does this \textit{Gaussian equivalence property} hold, and when does it break down?

This section presents a series of numerical experiments designed to answer these questions. We treat the ODEs derived under Gaussian assumptions (i.e. continuous-time limit of SGD under the Gaussian assumption) as a baseline prediction and compare them against direct SGD simulations on various structured, non-Gaussian inputs. Our investigation unfolds in a structured manner:
\begin{itemize}
    \item First, we demonstrate that without any intervention, learning dynamics with non-Gaussian inputs indeed diverge significantly from the theoretical predictions (Sec.~\ref{ssec:unstandardized}).
    \item Next, we test our central hypothesis: that a simple standardization of the inputs is a surprisingly powerful mechanism for recovering Gaussian-like dynamics, even for complex distributions. This finding holds a key implication that the learning process is more sensitive to low-order moments than the fine-grained details of the input distribution (Sec.~\ref{ssec:standardized_gm} and \ref{ssec:various_dist}).
    \item Finally, we explore the boundaries of this universality by introducing correlations, identifying the conditions under which the convergence breaks down and providing insight into the theory's limits of applicability (Sec.~\ref{ssec:block_dependent}).
\end{itemize}

For clarity, $Z$ denotes a standard normal random variable, $Z\sim\mathcal{N}(0,1)$, $C^{\mathcal{M}}$ denotes a random variable from the dimension-wise Gaussian-mixture setting and $C^{\mathcal{B}}$ denotes a random variable from the block-dependent Gaussian-mixture setting.

Additionally, we use $\overline{X}$ to indicate dimension wise standardized inputs. For example, for the block-dependent mixture, $\cbarBm$ is defined by 
\begin{equation}
\cbarBm_r \defeq \frac{C^{\mathcal{B}}_r - \mu_r}{\sigma_r}, \quad r=1,\cdots,D 
\end{equation}
where $\mu_r \defeq \E[C^{\mathcal{B}}_r]$ and $\sigma_r^2 \defeq \Var(C^{\mathcal{B}}_r)$. See Fig.~\ref{figure1} for the overall scheme used to compare ODE and SGD dynamics.
To account for the stochastic nature of SGD dynamics, which depends on the selection of samples at each update step, we run SGD multiple times from the same initial conditions (i.e., keeping the teacher model, student model, and feature matrix $F$ fixed), while varying the random seed for sample selection. The resulting dynamics are then averaged to obtain a more stable representation.

\subsection{Unstandardized Gaussian Mixture Results}\label{ssec:unstandardized}

We first assess how the dynamics change when inputs are drawn from unstandardized Gaussian mixtures, i.e., ${C}^\mathcal{M}$. We vary the mixture parameters by setting $\alpha\in\{0.01,0.1,1\}$, fix $\beta=10$, and consider $q\in\{2,8\}$ mixture components.
As observed in Fig.~\ref{figure3_1}, the dynamics under unstandardized Gaussian mixture settings significantly diverge from those under a simple Gaussian distribution. Although both the ODE dynamics and SGD dynamics start from identical initial conditions, significant discrepancies emerge over time. \\
Even for an arbitrary distribution, the dynamics of the second layer's weight $v$ (Eq.~(\ref{eqn:v})) maintain the same form but differ in the calculation of the function correlation. By tracking the difference in $v$ between ODE and SGD dynamics, we can assess the effect of distribution differences.\\
In Fig.~\ref{figure3_1}(d), we can see how the difference in $v$ deviates over time; as expected, the deviation becomes more distinct as time evolves. The deviation is small when $\alpha$ is sufficiently small because the means $\mu_i$ for the Gaussian mixture are mostly sampled at $0$, making it difficult to distinguish between samples from the Gaussian mixture and simple Gaussian distribution. 

\subsection{Standardized Gaussian Mixture Results}\label{ssec:standardized_gm}

We next analyze dynamics with standardized teacher inputs, $\overline{C}^{\mathcal M}$. For visualization, we vary the mixture parameters as before.
Analysis of the dynamics with standardized Gaussian mixtures yielded notable outcomes. As shown in Fig.~\ref{fig:std_dynamics}, these mixtures do not introduce any significant discrepancies between ODE dynamics and SGD dynamics.\\
These findings indicate that input standardization is the key factor underlying the observed convergence between ODE predictions and SGD dynamics.

\subsection{Various Distribution Setting Results}\label{ssec:various_dist}

Because Gaussian mixtures can flexibly approximate diverse distributions, we also examine SGD dynamics under several non-Gaussian laws. For standardization, the empirical mean and variance are estimated from the samples and used to form standardized inputs. The distributional parameters are summarized in Table~\ref{tab:dist}.

\begin{table}[htb]
\caption{Parameters for various distributions used in the experiments}
\begin{tabular}{l|c}
Distribution & Parameter(s)            \\ \hline \hline
uniform      & $a = 0$, $b=10$           \\
beta (1)     & $\alpha=0.5$, $\beta=0.5$ \\
beta (2)     & $\alpha=5$, $\beta=1$ \\
Poisson     & $\lambda=2$ \\
Laplace     & $\mu=0$, $b=1$ \\
Pareto     & $\alpha=5$ \\
Lorentz     & $x_0=0$, $\gamma=1$ \\ \hline
Gaussian mixture    & \begin{tabular}[c]{@{}c@{}}$p_i = 0.3, 0.7$,\\ $\mu_i \sim \mathcal{U}[-2,2]$, $\sigma_i \sim \mathcal{U}[0.5, 5]$\end{tabular}
\end{tabular}
\label{tab:dist}
\end{table}

In Fig.~\ref{fig:dynamics_dist}, we observe the expected convergence behavior across these distributions, with the exception of the Lorentz (Cauchy) case, which lacks finite mean and variance. This pattern suggests that homogeneity of low-order cumulants—specifically, the existence and alignment of the first two moments, is a key theoretical ingredient.

\subsection{Block-dependent Gaussian Mixture Results}\label{ssec:block_dependent}
We next examine standardized block-dependent Gaussian-mixture inputs, $\cbarBm$. Intuitively, the maximum-dependence parameter $m$ may affect the learning dynamics. To isolate this effect, we fix a large dimension, $N=D=4096$, and vary $m\in\{2^0,2^1,\ldots,2^6\}$.
Because the pair $(N,D)$ and all model components (teacher, student, and feature matrix) are held fixed across runs, any changes in the trajectories can be attributed to $m$.

Fig.~\ref{fig:figure_m} shows that the convergence behavior of the dynamics changes systematically with $m$. When $m$ is sufficiently small relative to $D$, the block dependence is weak enough that the standardized inputs behave effectively Gaussian, and the ODE and SGD dynamics coincide. In our setting the dependence has no preferred direction, so the degree to which the Gaussian limit dynamics are preserved decreases with the maximum magnitude of random correlations, as controlled by $m$. Because convergence is observed whenever $m\ll D$, these results suggest the presence of an $m/D$-type bound that governs dynamical convergence.

\section{Dynamics Convergence under block-dependent distributions}\label{Derivation}

The preceding section provided compelling empirical evidence for a remarkable phenomenon. The learning dynamics under a standardized, non-Gaussian inputs converge to a universal trajectory predicted by Gaussian theory. Furthermore, we observed that this universality gracefully degrades as input correlations, controlled by the ratio $m/D$, are introduced. This section provides the mathematical underpinnings for these empirical findings. Our central argument is that the observed universality in the dynamics is a direct consequence of universality in the underlying preactivation distribution $\{\lambda, \nu\}$.

To establish this, our analysis proceeds in two key steps. We first show that the preactivations become asymptotically Gaussian due to a central-limit-like effect on the weakly correlated blocks; second, we prove that their covariance matrix remains consistent with the Gaussian case. Together, these pillars provide a rigorous explanation for the convergence observed in our simulations and reveal why the $m/D$ ratio naturally emerges as the controlling factor in our theoretical bounds.

\subsection{Asymptotic Gaussianity under block-dependent distributions}
We establish asymptotic Gaussianity via a blockwise decomposition. 
For a fixed feature matrix $F \in \R^{D \times N}$ and weight vector $\widetilde{W}_m \in \R^D$, $U_i$ and $\nu_m$ with standardized block-dependent Gaussian mixture input $\overline{C}^{\mathcal{B}}$ are defined as:
\begin{align}
U_i &\defeq \frac{1}{\sqrt{D}} \sum_{r=1}^D F_{r,i} \overline{C}^{\mathcal{B}}_r, \quad \text{for } i=1, \dots, N, \\
\nu_m &\defeq \frac{1}{\sqrt{D}} \sum_{r=1}^D \widetilde{W}_{m,r} \overline{C}^{\mathcal{B}}_r, \quad \text{for } m=1, \dots, M.
\end{align}

The key insight is that both $U_i$ and $\nu_m$ can be expressed as a sum over the independent blocks. For a fixed column index $i$, we define the contribution of each block $b$ to the statistic $U_i$ as below:
\begin{equation}
T_b^{(U)} \defeq \sum_{r \in \mathcal{B}_b} \frac{F_{r,i}}{\sqrt{D}} \overline{C}^{\mathcal{B}}_r.
\end{equation}

Since the data vectors  $\Cbar_{\mathcal{B}_b}=\{\overline{C}^{\mathcal{B}}_i\}_{i\in\mathcal{B}_b}$ are independent for different blocks $b$, the random variables $\{T_b^{(U)}\}_{b=1}^{n_b}$ are mutually independent. The total statistic is simply the sum of these independent contributions,
\begin{equation}
U_i = \sum_{b=1}^{n_b} T_b^{(U)}.
\end{equation}
The same decomposition applies to $\nu_m$ by defining
\begin{equation}
    T_b^{(\nu)} \defeq \sum_{r \in \mathcal{B}_b} (\widetilde{W}_{m,r}/\sqrt{D}) \Cbar_r.
\end{equation}
By block wise independence, we can utilize Berry--Esseen bounds.
\begin{theorem}[Uniform Berry--Esseen bounds]. The Berry--Esseen theorem \citet{berry1} and \citet{berry2}; see, e.g., \citet{berry3}, for a sum of independent, non-identically distributed, zero-mean random variables $X_b$ states that if $\sum_b \Var(X_b) = 1$, then 
\begin{equation}
\sup_{x \in \R} | P(\sum_b X_b \le x) - \Phi(x) | \le C_{BE} \sum_b \mathbb{E}[|X_b|^3],
\end{equation}
where $\Phi(x)$ is the standard normal CDF and $C_{BE}$ is a universal constant.
\end{theorem}
Apply this to the normalized sum $U_i/\sigma_{U,i}$, where $\sigma_{U,i}^2\defeq\Var(U_i)$. Let $Z_b\defeq T_b^{(U)}/\sigma_{U,i}$. Then ${Z_b}$ are independent with $\sum_b\Var(Z_b)=1$.
Under the following assumption,
\begin{assumption}[Block bound assumption]\label{main:assumption:energy2}
For some $\beta_U>0$
\begin{equation}
\Var(T_b^{(U)})\le\;\cfrac{C_\Sigma}{D}||F_b||^2 \leq C_E (m/D)^{\beta_U},
\end{equation}
\end{assumption}

each block's contribution scales with $m/D$, which implies

\begin{equation}
\sum_{b=1}^{n_b} \mathbb{E}[|Z_b|^3] = \frac{1}{\sigma_{U,i}^3}\sum_{b=1}^{n_b}\E|T_b^{(U)}|^3 \le  \Bigl({\frac{m}{D}}\Bigr)^{3\beta_U/2-1}.
\end{equation}

Intuitively, global scaling contributes the $1/D$ factor, while summation over blocks contributes the $m$ factor. Details are in Appendix~\ref{appendix:Gaussianity}.

In summary, below theorem holds.
\begin{theorem}
\label{thm:U_normality_BE}
There exists a constant $C_U$, independent of $D$ and $m$, such that for the Kolmogorov-Smirnov distance $d_{\mathrm{KS}}(\cdot, \cdot)$:
\begin{equation}
d_{\mathrm{KS}}\left( \frac{U_i}{\sigma_{U,i}}, \mathcal{N}(0,1) \right) \le C_U \Bigl({\frac{m}{D}}\Bigr)^{3\beta_U/2-1}.
\end{equation}
where $\sigma_{U,i}^2=\Var(U_i)$.
An identical argument holds for $\nu_m$, yielding:
\begin{equation}
d_{\mathrm{KS}}\left( \frac{\nu_m}{\sigma_{\nu,m}}, \mathcal{N}(0,1) \right) \le C_\nu \Bigl({\frac{m}{D}}\Bigr)^{3\beta_\nu/2-1},
\end{equation}
where $\sigma_{\nu,m}^2 = \Var(\nu_m)$.
\end{theorem}
A similar conclusion holds for $\lambda_k$; see Appendix~\ref{appendix:Gaussianity} for details.

In addition, the Cram\'er--Wold theorem~\cite{cramer} implies that if every linear combination $a_1\lambda_1+\cdots+a_K\lambda_K$ is Gaussian, then $(\lambda_1,\ldots,\lambda_K)$ is jointly Gaussian.

\subsubsection{Numerical verification}

\begin{figure}[htb]
\centerline{\includegraphics[width=0.5\textwidth]{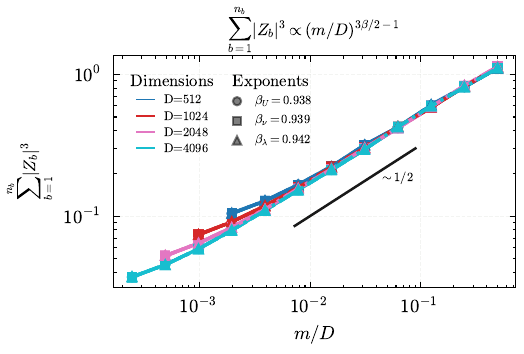}}
\caption{For each statistic ($U$, $\nu$, $\lambda$) we form the normalized block terms $Z_b$ (e.g., $Z_b=T_b^{(U)}/\sigma_{U,i}$ for $U$; analogously for $\nu$ and $\lambda$) and plot $\sum_{b=1}^{n_b}\mathbb{E}|Z_b|^3$ versus $m/D$ on log-log axes. Across different ambient dimensions $D$, the curves collapse onto a single line with slope close to $1/2$, consistent with the theory $\sum_b \mathbb{E}|Z_b|^3 \lesssim (m/D)^{3\beta/2-1}$ and $\beta=1$ in the idealized design. The lower the transparency, the larger the sample size, since $P=10^3$ already sufficient for deducing numerical error in current aspect, all different sample size results are collapsed. See Appendix~\ref{appendix:Gaussianity} for definitions.}
\label{fig:figure_3rd}
\end{figure}

\begin{figure}[htb]
\centerline{\includegraphics[width=0.5\textwidth]{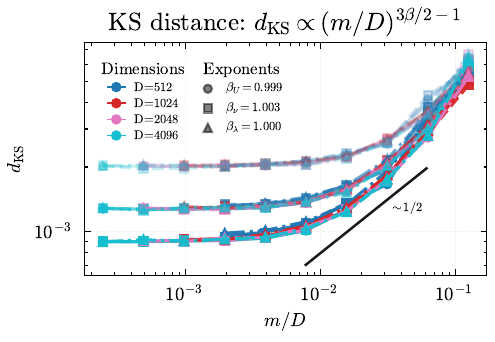}}
\caption{Kolmogorov-Smirnov distance $d_{\mathrm{KS}}$ between the standardized statistics (e.g., $U_i/\sigma_{U,i}$, $\nu_m/\sigma_{\nu,m}$, $\lambda_k/\sigma_{\lambda_k}$) and $\mathcal N(0,1)$, plotted against $m/D$ on log-log axes. In the mid-range of $m/D$ the curves collapse with slope close to $1/2$, matching the Berry--Esseen prediction $d_{\mathrm{KS}}\lesssim \sum_b \mathbb{E}|Z_b|^3 \lesssim (m/D)^{3\beta/2-1}$ with $\beta=1$. The lower the transparency, the larger the sample size, we can see that as large sample size make more clear collapse to $\beta \to 1$ in mid range.}
\label{fig:figure_ks}
\end{figure}

Closed form exponents are difficult in the general block-dependent Gaussian mixture model, so we verify the bounds under a simplified, idealized design. $F$ are drawn i.i.d. Gaussian and then column-wise normalized so that $\|F_{\cdot,i}\|^2=D$, and blocks are uniform with $n_b=D/m$. In this regime, the exponent from block contribution moment specializes to $\beta=1$; therefore
\begin{equation}\label{eq:num-3rd}
\sum_{b=1}^{n_b}\mathbb{E}|Z_b|^3 \ \le\ C_3 \Big(\frac{m}{D}\Big)^{\frac{3\beta}{2}-1}
\ =\ C_3 \Big(\frac{m}{D}\Big)^{\frac{1}{2}},
\end{equation}
and, by Berry--Esseen,
\begin{equation}\label{eq:num-ks}
d_{\mathrm{KS}}\Big(\frac{U_i}{\sigma_{U,i}},\,\mathcal N(0,1)\Big)
\ \le\ C_{\mathrm{BE}} \sum_{b=1}^{n_b}\mathbb{E}|Z_b|^3
\ \le\ C\,\Big(\frac{m}{D}\Big)^{\frac{1}{2}}.
\end{equation}
Analogous statements hold for $\nu$ and, after the remainder control in the projection argument, for $\lambda$. See Appendix~\ref{appendix:GaussianLimit} for details.

To emulate the asymptotic regime, we choose large sample sizes $P$ and dimensions $D$. 
For the third-moment measurements we set $P=\{10^3, 2.5\times10^3, 5\times10^3 \}$ and take $D\in\{1024,\,2048\}$. 
For the $d_{\mathrm{KS}}$ measurements we set $P=\{10^5, 2.5\times10^5, 5\times10^5 \}$ and take $D\in\{512,\,1024,\,2048,\,4096\}$, because the empirical $d_{\mathrm{KS}}$ error has a non-negligible sampling floor and a larger $P$ is needed to reveal the $(m/D)^{1/2}$ decay. 
We also sweep $m$ so that $m$ and $D$ are both large while the ratio $m/D$ stays finite, which stabilizes the identification $\beta\to 1$. 
At the same time we avoid taking $D/m$ too small, so that the number of blocks remains sufficiently large for averaging to be effective.

Figures~\ref{fig:figure_3rd}-\ref{fig:figure_ks} show clear data collapse across $D$ when plotted against $m/D$. 
The third-moment curves exhibit a slope $\simeq 1/2$ over the entire range, in agreement with~\eqref{eq:num-3rd}. 
For $d_{\mathrm{KS}}$, the slope $\simeq 1/2$ is most visible in the mid-range of $m/D$. 
This aligns with the fact that the numerical error in estimating $d_{\mathrm{KS}}$ decreases when the number of independent blocks $D/m$ is large enough to average out fluctuations, and when $m$ is large so that, with high probability, effective exponent satisfies $\beta \approx 1$.

These observations support the Gaussianity predictions in the thermodynamic limit.

\subsection{Function Correlation Decomposition}
By the blockwise Berry--Esseen bound, the preactivations ${\lambda,\nu}$ are asymptotically jointly Gaussian. To proceed, it remains to verify that their covariance retains the same form after replacing $C$ by standardized block-dependent mixture inputs.
Since the sum of Gaussian variables remains Gaussian even if they are correlated, we conclude that $\{\lambda, \nu\}$ follows a joint Gaussian distribution.

We approximate function correlations of a random vector $S$ by those of its Gaussian counterpart as follows. 
Let $S\in\R^p$ be any statistic that decomposes as a sum of independent, mean-zero block contributions,
\begin{equation}\label{main:eq:S-sum}
  S \;=\; \sum_{b=1}^{n_b} W_b,
  \quad
  W_b \;=\; \Phi_b\!\big(\CbarBm_{\mathcal B_b}\big)\;-\;\E\Big[\Phi_b\!\big(\CbarBm_{\mathcal B_b}\big)\Big]
  ,
\end{equation}
where each $\Phi_b:\R^{|\mathcal B_b|}\to\R^p$ is a map depending only on the block coordinates
$\CbarBm_{\mathcal B_b}$. We control the second-order size of each contribution by following assumption

\begin{assumption}[Block bound assumption]\label{main:assumption:energy}
\begin{equation}
    \E\|W_b\|^2 \;\le\; (m/D)^{\beta}.
\end{equation}
\end{assumption}
Under this assumption, the expectation gap $\E h(S)-\E h(Z)$ decomposes into two error channels: a conditional-variance term of order $(m/D)^{\beta-1/2}$ and a cubic-moment term of order  $(m/D)^{3\beta/2-1}$. Taking the dominant contribution yields the exponent $\gamma=\min\{\beta-\tfrac12,\tfrac{3\beta}{2}-1\}$, and the following lemma holds.

\begin{lemma}[Smooth-test approximation~\cite{exchange}]\label{main:lem:smooth-test}
For $S$ as in \eqref{main:eq:S-sum} and $Z\sim\mathcal N(0,\Sigma)$ with $\Sigma=\Cov(S)$, under \eqref{eq:block-spectral-bds} and Assumption~\ref{main:assumption:energy}, for any $h\in C^3(\R^p)$ with bounded derivatives,
\begin{equation}\label{eq:smooth-rate-detailed}
\big|\E h(S)-\E h(Z)\big|
\;\le\; C_h\left({\frac{m}{D}}\right)^\gamma.
\end{equation}
\end{lemma}

Thus, function correlations under $S$ can be approximated by those under the Gaussian $Z$ up to an error of order $(m/D)^{\gamma}$. Full details are provided in Appendix~\ref{Appendix:FunctionDecompose}.

As an illustration, consider $Z_\rho=(X_\rho,Y_\rho)$ with mean zero and covariance
\begin{equation}
\Sigma(\rho)=\begin{pmatrix}\Sigma_{xx} & \rho \Sigma_{xy}\\\rho \Sigma_{xy}^\top & \Sigma_{yy}\end{pmatrix}.
\end{equation}
so the cross-covariance is controlled by $\rho$. For $h(\rho)\defeq \E[\varphi(Z_\rho)]$ with $\varphi(x,y)=f(x)g(y)$, Price’s theorem~\cite{price} yields, 
\begin{equation}
    \frac{d}{d\rho}\,\mathbb E[\varphi(Z_\rho)]=\frac12\,\mathbb E\!\big[\tr(\nabla^2\varphi(Z_\rho)^\top\Sigma'(\rho)\big)], 
\end{equation}
with
\begin{equation}
    \Sigma'(\rho)=\begin{pmatrix}0&\Sigma_{xy}\\\Sigma_{yx}&0\end{pmatrix}
\end{equation}
giving
\begin{equation}\label{eq:price-first-deriv}
    h'(\rho)=\mathbb E\!\big[\nabla f(X_\rho)^\top\,\Sigma_{xy}\,\nabla g(Y_\rho)\big].
\end{equation}
For Gaussian $X_0\sim\mathcal{N}(0,\Sigma_{xx})$, $Y_0\sim\mathcal{N}(0,\Sigma_{yy})$, Stein's Lemma says that,
\begin{equation}\label{eq:stein-grad}
\begin{aligned}
     \mathbb E[\nabla f(X_0)]&=\Sigma_{xx}^{-1}\mathbb E[(X_0)f(X_0)],\\
     \mathbb E[\nabla g(Y_0)]&=\Sigma_{yy}^{-1}\mathbb E[(Y_0)g(Y_0)],   
\end{aligned}
\end{equation}
Plugging in to
\begin{equation}
    h(\rho)=h(0)+\rho\,h'(0)+O(\rho^2)
\end{equation}
This results to 
\begin{equation}
    h(\rho) = \rho \E[X_0f(X_0)]^T\cfrac{\Sigma_{xy}}{\Sigma_{xx}\Sigma_{xy}} \E[Y_0g(Y_0)] + \mathcal{O}(\rho^2).
\end{equation}
Under the stated boundedness conditions, $\E[\varphi(S)]$ can be approximated by its Gaussian counterpart $\E[\varphi(Z)]$.

\subsection{Covariance Consistency}
Consider $U_i$ formed from standardized block-dependent Gaussian mixture inputs $\cbarBm$. 
Let $U=(U_i,U_j)$ with mean $\mu=(\mu_i,\mu_j)$ and covariance $\Sigma$, and
let $Z\sim\mathcal N(\mu,\Sigma)$.

Define the centered pairs $\tilde U=(U_i-\mu_i,U_j-\mu_j)$ and
$\tilde Z=(Z_i-\mu_i,Z_j-\mu_j)$, so that $\E\tilde U=\E\tilde Z=0$ and
$\Cov(\tilde U)=\Cov(\tilde Z)=\Sigma$.
For any $\psi\in C^3(\R^2)$ with polynomially bounded derivatives, set
\begin{equation}
\psi_\mu(u,v)\defeq \psi(u+\mu_i,\,v+\mu_j).
\end{equation}
Then
\begin{equation}
\E\psi(U_i,U_j)-\E\psi(Z_i,Z_j)
= \E\psi_\mu(\tilde U_i,\tilde U_j)-\E\psi_\mu(\tilde Z_i,\tilde Z_j).
\end{equation}
Applying Lemma~\ref{main:lem:smooth-test} yields 
\begin{equation}\label{eq:mean-shift-smooth}
\big|\E\psi(U_i,U_j)-\E\psi(Z_i,Z_j)\big|
\ \le\
C_\psi(m/D)^{\gamma'},
\end{equation}
for some exponent $\gamma'$ and constant $C_\psi$.
Taking $\psi(u,v)=f(u)$ and $\psi(u,v)=f(u)f(v)$, we obtain
\begin{equation}
    |\Cov\!\big(f(U_i),f(U_j)\big)-
\Cov\!\big(f(Z_i),f(Z_j)\big)| \leq (m/D)^{\gamma'}.
\end{equation}
For the Gaussian counterparts, define
\begin{equation}
    \mu_i\defeq \E[Z_i],\quad \sigma_{ij}^2\defeq \Cov(Z_i,Z_j)
\end{equation}
\begin{equation}
a_i\defeq \E[f(Z_i)],\quad b_i\defeq \E[Z_i f(Z_i)].
\end{equation}
Price’s theorem then gives
\begin{equation}\label{main:euiuj}
\begin{aligned}
    \E[f(Z_i)f(Z_j)] = a_ia_j + \cfrac{(b_i-\mu_ia_i)\sigma_{ij}^2(b_j-\mu_ja_j)}{\sigma_{{ii}}^2\sigma_{{jj}}^2}
\end{aligned}
\end{equation}

By definition, under standardized block-dependent Gaussian mixture inputs,
\begin{equation}
Q_{k\ell}\equiv \E\!\big[\overline\lambda_k\,\overline\lambda_\ell\big]
\end{equation}
is determined by
\begin{equation}
    \Cov\!\big(f(U_i),f(U_j)\big),
\end{equation}
and the Gaussian counterparts be
\begin{equation}
    \mathbb{E}[f(Z)f(Z)] = a^2 + \frac{1}{D}\sum^D_{r=1} F_{r,i}F_{r,j}b^2
\end{equation}
where $a\defeq \E[f(Z)]$ and $b\defeq \E[Z f(Z)]$ for $Z\sim\mathcal N(0,1)$. Consequently, the difference between the mixture and Gaussian evaluations of $\E[f(U_i)f(U_j)]$ vanishes in the thermodynamic limit, and the expression for $Q$ coincides with that in Appendix~\ref{appendix:Q}. A similar argument shows that $R$ retains the form in Eq.~(\ref{eqn:R}) via the approximation of $\E[\Cbar^{\mathcal B}_r f(U_i)]$. Full details are provided in Appendix~\ref{appendix:GaussianLimit}.

\subsubsection{Numerical verification}
\begin{figure}[htb]
\centerline{\includegraphics[width=0.5\textwidth]{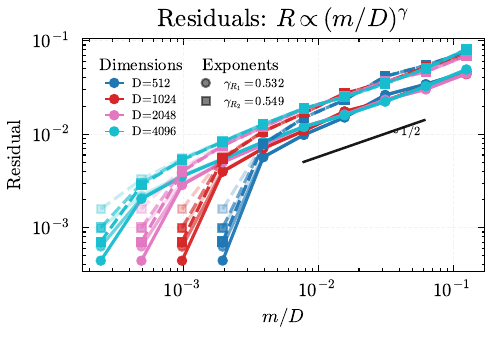}}
\caption{Numerical verification of residual scaling. The residuals $R_1\propto(m/D)^{\gamma}$ (Eq.~\eqref{main:euiuj}) and $R_2\propto(m/D)^{\gamma}$ (Eq.~\eqref{main:ecrui}) are plotted against $m/D$ for various system dimensions $D$. Curves collapse across $D$, confirming the predicted $(m/D)^{\gamma}$ law with $\gamma=\min\{\beta-\tfrac12,\tfrac{3\beta}{2}-1\}$. The lower the transparency, the
larger the sample size. The results shows clear collapse to $\gamma\to1/2$.}
\label{fig:figure_res}
\end{figure}

Because the exponent governing the bound is difficult to obtain in closed form for the blockwise mixture, we estimate it empirically (see Fig.~\ref{fig:figure_res}). Specifically, we compute the residuals
\begin{equation}\label{main:euiuj}
R_1 = \left|\mathbb{E}[f(U_i)f(U_j)] - \left( a^2 + \frac{1}{D}\sum^D_{r=1} F_{r,i}F_{r,j}b^2\right)\right| 
\end{equation}
which are hypothesized to satisfy
\begin{equation}
R_1    <  C\left({\frac{m}{D}}\right)^\gamma + \mathcal{O}(\rho^2)
\end{equation}
and
\begin{equation}\label{main:ecrui}
R_2 = \left|\mathbb{E}[f(U_i)\cbarBm_r] - \left( \frac{1}{\sqrt{D}}F_{r,i} \mathbb{E}[U_i f(U_i)]\right)\right|
\end{equation}
which are hypothesized to satisfy
\begin{equation}
R_2     <  C\left({\frac{m}{D}}\right)^\gamma + \mathcal{O}(\rho^2)
\end{equation}
where $a\defeq\E[f(Z)]$ and $b\defeq\E[Zf(Z)]$ for $Z\sim\mathcal N(0,1)$, and $\rho$ quantifies the weak cross-correlation in the Gaussian proxy.
Under a Gaussian feature matrix with column normalization $\|F_{\cdot,i}\|^2=D$ and uniform blocks $n_b=D/m$, the effective exponent satisfies $\beta\to 1$ with high probability, hence $\gamma\to \tfrac12$. 

To emulate the asymptotic regime, we set $P=\{10^5, 2.5\times10^5, 5\times10^5 \}$ and take $D\in\{512,\,1024,\,2048,\,4096\}$. 
We sweep $m$ so that $m$ and $D$ are both large while the ratio $m/D$ remains finite, which stabilizes $\beta\approx 1$. 
We plot $R_1$ and $R_2$ against $m/D$ on log-log axes and fit the slope in the large-$m$ regime. 
In Fig.~\ref{fig:figure_res}, the curves collapse across $D$ under this rescaling, and the dominant error exhibits an $(m/D)^{\gamma}$ law with $\gamma$ close to $1/2$, consistent with $\gamma\approx\tfrac12$ and supporting covariance consistency in the thermodynamic limit.

\subsection{Summary of Theoretical Results.}
Our mathematical analysis provides a firm theoretical foundation for the phenomena observed in Section~\ref{Results}. We have established two key properties for the preactivation distribution $\{\lambda, \nu\}$ under standardized, block-dependent Gaussian mixture inputs in the thermodynamic limit:
\begin{enumerate}
    \item \textbf{Asymptotic Gaussianity:} The joint distribution of preactivations converges to a multivariate Gaussian. The rate of this convergence is explicitly controlled by the relative correlation strength, $m/D$.
    \item \textbf{Covariance Consistency:} The mean and covariance matrix of this limiting Gaussian distribution are converges to those derived under simple Gaussian inputs.
\end{enumerate}
Together, these results rigorously explain why the learning dynamics exhibit universality. The dynamics remain unchanged because the underlying statistical properties of the preactivations, which drive the learning process, are themselves universal.

\section{Discussion}\label{Discussion}
We discuss limitations and potential extensions of our results. Where closed-form analysis is out of reach, we complement the discussion with targeted simulations. First, we identify edge cases in which convergence is observed empirically but the ODE description does not apply. Second, we examine the discrepancy between ODE and SGD dynamics when input correlations exceed the weak-correlation regime; although our theory does not quantify this error, we assess it empirically.

\subsection{Results under affine transformations of Gaussian inputs}

\begin{figure*}[htb]
\centerline{\includegraphics[width=\textwidth]{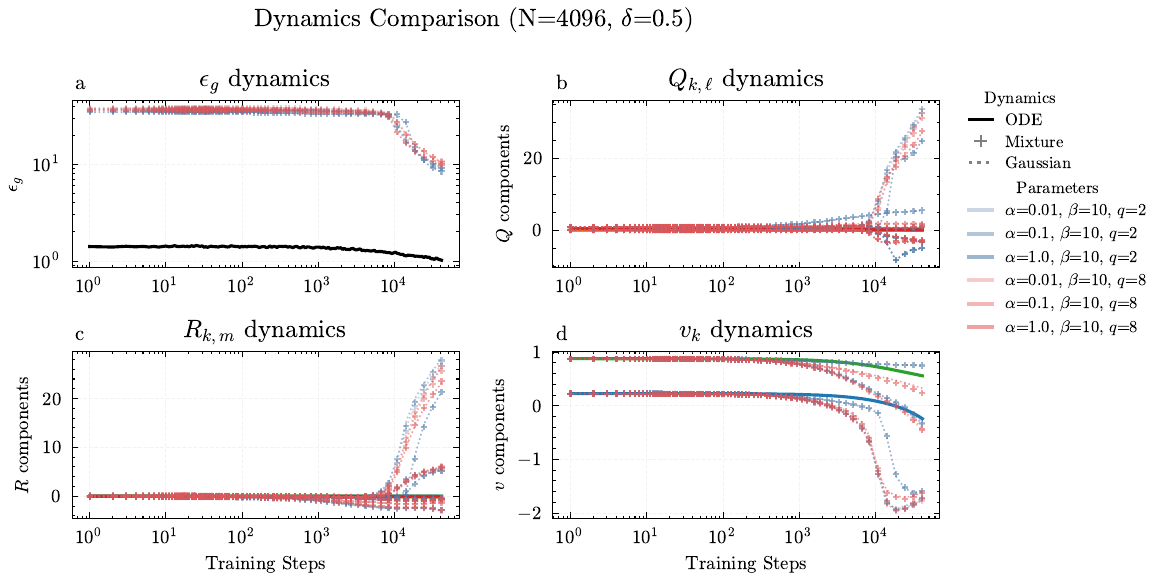}}
\caption{Examples of dynamics under Gaussian mixtures and its affine Gaussian proxy $m=1,2,\ldots,64$, $N=D=4096$. Dynamics of (a) generalization error $\epsilon_g$, (b) covariance matrix $Q$, (c) covariance matrix $R$, and (d) weight of the second layer $v$. The SGD results are averaged over 5 runs. We can see the convergence between SGD under Gaussian mixture and SGD under Gaussian proxy not but ODE}
\label{fig:figure_affine}
\end{figure*}
One might ask whether convergence requires identity of the means and covariances in all dimension.
For jointly Gaussian variables, distributional equality is determined by Gaussianity together with matching first and second moments; thus heterogeneity in the covariance (e.g., the case $a_i\neq a_j$ in \eqref{main:euiuj}) warrants attention.
Convergence in distribution is preserved under affine transformations. Hence, if the manifold inputs are unstandardized, as in a Gaussian mixture $C^{\mathcal M}$ with coordinatewise mean and variance $(\mu_r,\sigma_r^2)$, we may consider the affine proxy
\begin{equation}
Z_r^{\mathcal M}\;\defeq\;\mu_r+\sigma_r Z_r,\quad Z_r\sim\mathcal{N}(0,1).
\end{equation}
Under weak correlation and in the thermodynamic limit, this affine Gaussian proxy and the Gaussian mixture induce the same preactivation means and covariances; moreover, $(\nu,\lambda)$ remain Gaussian after affine transformations. Consequently, the dynamics generated by an affine-transformed Gaussian input and by the corresponding Gaussian mixture are dynamically equivalent. Fig~\ref{fig:figure_affine} shows that, even when the empirical dynamics coincide (same means and variances), a full analytical tracking can fail: heterogeneous, dimension-dependent variances hinder the separation needed to express the covariance through well-decoupled projected matrices (cf. \eqref{append:decoupling}). This identifies a genuine limitation of the ODE approach in strongly heterogeneous settings.

\subsection{Correlation Effects on Dynamic Behavior}

\begin{figure*}[htb]
    \centering
    \includegraphics[width=\linewidth]{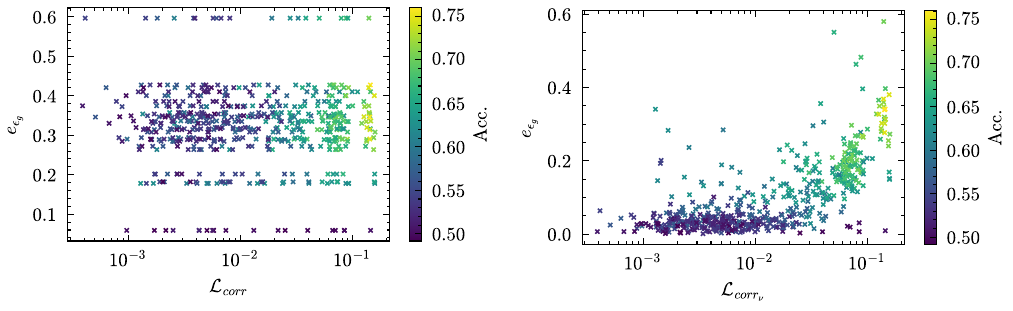}
    \caption{Left panel: Relationship between total correlation loss $\mathcal{L}_{corr}$ (x-axis) and dynamic error $e_{\epsilon_g}$ (y-axis). The color represents classification accuracy on the training dataset, and each point corresponds to a specific teacher model and teacher model input pair obtained from the encoder-teacher training process. Right panel: The same plot, but with the x-axis replaced by $\mathcal{L}_{corr_{\nu}}$, isolating the correlation effect of $\nu$. This results demonstrates the correlation between the teacher weights and inputs as the dominant factor driving the deviation from universal dynamics.}
    \label{fig:correlation}
\end{figure*}

\begin{figure}[htb]
    \centering
    \includegraphics[width=\linewidth]{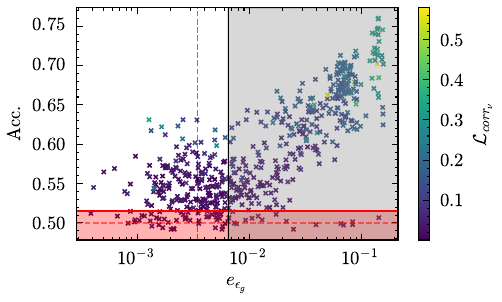}
    \caption{Revised visualization highlighting the relationship among dynamic error $e_{\epsilon_g}$ (x-axis), accuracy (y-axis) and $\mathcal{L}_{corr_{\nu}}$ (color). The red dotted and solid horizontal lines indicate the mean accuracy, $0.5$ and $0.5+3\sigma_{\text{acc}}$ of a random classifier, respectively. The black dotted and solid lines represent the empirical stochastic error magnitude, calculated as the mean $e_{\epsilon_g}$, $e_{\text{base}}$ and $e_{\text{base}}+\sigma_{\text{base}}$, respectively. The key finding is the existence of a population of models in the first (top-left) quadrant. They are both informative and their dynamics are well approximated, demonstrating a possible practical sweet spot for its application.}
    \label{fig:correlationw/quad}
\end{figure}

\begin{figure*}
    \centering
    \includegraphics[width=\textwidth]{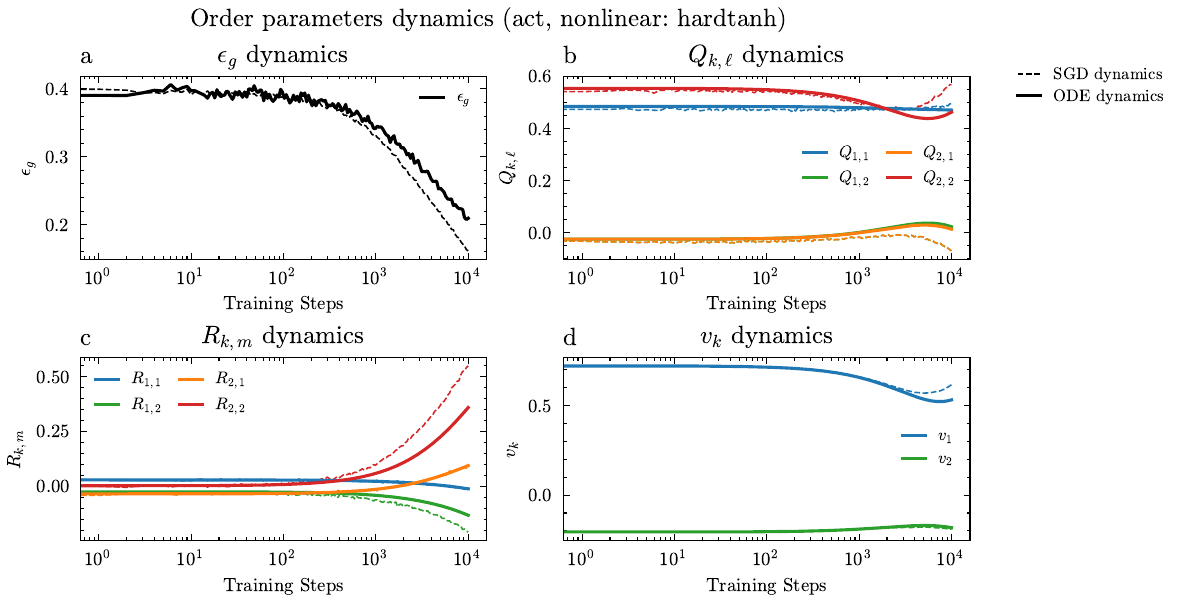}
    \caption{Example dynamics for selected teacher model and teacher model input.This figure shows the learning dynamics for a single model selected from the top-left quadrant of Fig.~\ref{fig:correlationw/quad}. Dynamics of (a) generalization error $\epsilon_g$, (b) covariance matrix $Q$, (c) covariance matrix $R$, and (d) weight of the second layer $v$. The SGD results are averaged over 5 runs. This provides a practical example suggesting our framework's potential utility in moderately correlated, real-world-inspired scenarios.}
    \label{fig:dynamics_mnist}
\end{figure*}

As demonstrated in our analysis, the dynamics continue to converge in the presence of certain correlations, as shown for $\cbarBm$, where effective correlations diminish in the thermodynamic limit. This property arises due to our choice of teacher model input and feature matrix $F$, both of which are sampled from their respective distributions, as well as the teacher and student model weights, which are drawn i.i.d. from a standard Gaussian distribution.
In theoretical approaches, assuming uncorrelated random variables is a common choice. For instance, in both the hidden manifold model and the conventional teacher-student model, the teacher model itself is constructed using i.i.d. weights, ensuring no intrinsic correlation with its input. However, in practical applications, correlations naturally emerge. If the teacher model is \textit{informative}, meaning it encodes meaningful structure from the data, correlations between its weights and inputs become inevitable. 
This trade-off between informativeness and weak correlation poses a challenge in applying the teacher-student framework to real-world scenarios. 

Moreover, these correlations do not only arise between teacher model weights and inputs but also among the teacher model weights themselves and among inputs themselves.
In this subsection, we empirically investigate this trade-off using real-world data. Since our mathematical justification of convergence depends on the weak correlation assumption, we cannot theoretically determine how increasing correlation affects dynamics, but we explore this question through empirical analysis.
\subsubsection{Correlated setting with real-world data}
To study weakly to moderately correlated settings and assess applicability to real-world data, we employ an encoder-teacher model framework. In this setup:
\begin{itemize}
    \item An encoder model is trained to encode the MNIST dataset into a lower-dimensional representation, $\mathcal{C}$.
    \item A teacher model receives the standardized version of the encoded output, represented as $\overline{\mathcal{C}}$, and is trained to classify the original MNIST images as odd or even.
\end{itemize}
Since Gaussian mixtures can approximate arbitrary distributions in the large-component limit, we continue to use the notation $\mathcal{C}$ in this setting. If the encoder-teacher model is well trained, the standardized input $\overline{\mathcal{C}}$ to the teacher model retains compressed information from MNIST, and the teacher model itself contains structured information necessary for classification.
Additionally, to explicitly control correlation effects, we modify the training objective of the teacher model by incorporating a correlation regularization loss:
\begin{equation}
    \mathcal{L} = \lambda_{corr}[\mathcal{L}_{corr,{\widetilde{W}}}+\mathcal{L}_{corr,U}+\mathcal{L}_{corr,\nu}] + |y-\hat{y}|^2 
\end{equation}
where 
\begin{align}
    \mathcal{L}_{corr,\nu} &= |\operatorname{Cov}(\nu_i,\nu_j|\overline{\mathcal{C}})-\operatorname{Cov}(\nu_i,\nu_j|Z)|\\
    \mathcal{L}_{corr,U} &= |\operatorname{Cov}(U_i,U_j|\overline{\mathcal{C}})-\operatorname{Cov}(U_i,U_j|Z)|\\
    \mathcal{L}_{corr,{\widetilde{W}}} &= |\operatorname{Cov}(\widetilde{W}_i,\widetilde{W}_j|\overline{\mathcal{C}})-\operatorname{Cov}(\widetilde{W}_i,\widetilde{W}_j|Z)|.
\end{align}
Here, $\operatorname{Cov}(a,b|X)$ represents the covariance of random variables $a$ and $b$ under teacher model input distribution $X$, and $Z$ is normal Gaussian distribution. $|\cdot|$ denote element-wise $\ell_1$ norm. By adjusting the strength of the correlation loss $\lambda_{corr}$, we obtain teacher models and inputs $\overline{\mathcal{C}}$ with varying degrees of informativeness and correlation.
We train the encoder-teacher model using $10^4$ samples from MNIST, reducing the input dimension to $D=500$, $\mathcal{C}\in\R^D$, lower than the original MNIST dimension $28\times28$. Training is performed with full-batch Adam optimization with learning rate $0.001$, and we terminate training after 30 epochs. We vary $\lambda_{corr}$ over the range $[1,10]$.

\subsubsection{Empirical causes of dynamic error}
Using the trained teacher models and inputs $\mathcal{C}$, we measure:
\begin{itemize}
    \item Label accuracy ($\operatorname{Acc.}$), which reflects the degree of informativeness of the teacher model.
    \item Total correlation loss ($\mathcal{L}_{corr}$), which quantifies the correlation between internal variables.
    \item Dynamic error ($e_{\mathcal{D}}$), which measures the discrepancy between ODE and SGD dynamics:
    \begin{equation}
    e_\mathcal{D} = ||\mathcal{D}_{\operatorname{SGD}}(\tau)-\mathcal{D}_{\operatorname{ODE}}(\tau)||_2
    \end{equation}
\end{itemize}
where $\mathcal{D}$ represents variables such as $\epsilon_g$, $Q_{k,\ell}$, $R_{k,m}$ or $v_k$, and $\tau=1000$ denotes the evaluation time step. The $\ell_2$ norm $||\cdot||_2$ quantifies the error magnitude.
Since SGD updates introduce randomness due to sample selection, we compute $e_\mathcal{D}$ by averaging over 20 runs of SGD with a fixed student model. We also evaluate the accuracy using training samples $10^4$. As each dynamic variable is updated proportional to the magnitude of each timestep error, which relate to $\epsilon_g$, we mainly focus on $\epsilon_g$ as a dynamic quantity $\mathcal{D}$.

Fig.~\ref{fig:correlation} illustrates the relationships among correlation loss, accuracy, and dynamic error. As correlation loss increases, accuracy also increases, exhibiting a strong correlation. This aligns with our expectation that more informative models deviate further from weakly correlated dynamics. The Pearson correlation coefficients $\rho$ confirm this trend, given by $\rho(\operatorname{Acc.}, e_{\epsilon_g})=0.774$.
However, we do not observe a direct correlation between total correlation loss $\mathcal{L}_{corr}$ and dynamic error $e_{\epsilon_g}$ or between $\mathcal{L}_{corr}$ and accuracy. To investigate further, we decompose the correlation loss into its components: $\mathcal{L}_{corr,\nu}$, $\mathcal{L}_{corr,U}$, and $\mathcal{L}_{corr,{\widetilde{W}}}$.
Unlike $\mathcal{L}_{corr,U}$ and $\mathcal{L}_{corr,{\widetilde{W}}}$, which have weak correlation with accuracy, we find that $\mathcal{L}_{corr,\nu}$ strongly correlates with both accuracy and dynamic error. 
In Fig.~\ref{fig:correlation}, the right panel directly show strong correlation among correlation loss, accuracy, and average dynamics error. A high correlation loss corresponds to high accuracy with a large dynamic error, whereas a small correlation loss corresponds to low accuracy and a small dynamic error, and vice versa. The Pearson correlation coefficients $\rho$ among these variables are given by $\rho(\mathcal{L}_{corr,\nu}, \operatorname{Acc})=0.833$, $\rho(\mathcal{L}_{corr,\nu}, e_{\epsilon_g})=0.762$, and $\rho(e_{\epsilon_g}, \operatorname{Acc})=0.774$.
From these observations, we hypothesize that `the correlation between teacher model inputs $\overline{\mathcal{C}}$ and teacher model weights $\widetilde{W}$ plays a crucial role in dynamic consistency.'
Theoretically, since our proofs assume weak correlation, we cannot rigorously quantify its detailed effect. However, empirically, we show correlation between teacher model and teacher model input appears to be the dominant factor.

\subsubsection{Possible extensions in correlated scenarios}
Given the trade-off between correlation strength and model informativeness, we now consider whether our theoretical approach can still provide insights for real-world applications.
To test this, we construct the hypothesis:
``We cannot track the dynamics of an informative model.''
To refute this hypothesis, we must show that even for informative models, dynamic tracking remains feasible.
First, a non-informative model can be approximated as a random classifier, which has a $50\%$ chance of correct classification. 
Given $n$ samples, the probability of correctly labeling $k$ samples follows a binomial distribution, leading to an accuracy variance of $\sigma_{\text{acc}} = 1/(2\sqrt{n})$. For our case $n=10^4$, a model achieving accuracy above $0.5 + 3\sigma_{\text{acc}}$ can be considered informative with high confidence.
Second, we assess whether we can track dynamics under moderate correlation. Even when the teacher model input is Gaussian, stochastic error in SGD introduces a baseline error between SGD and ODE dynamics. We denote the mean and standard deviation of this baseline error as $e_{\text{base}}$ and $\sigma_{\text{base}}$. If the dynamic error from a non-Gaussian input $\overline{\mathcal{C}}$ remains within $e_{\text{base}} + \sigma_{\text{base}}$, we conclude that tracking is feasible to some extent. So, we calcuate mean and variance from $20$ individual SGD dynamics under same initial conditions (fixed teacher, student model and feature matrix $F$).

Fig.~\ref{fig:correlationw/quad} visualizes relation between accuracy and dynamic error, with thresholds for informativeness ($0.5 + 3\sigma_{\text{acc}}$) and tracking feasibility ($e_{\text{base}} + \sigma_{\text{base}}$). The first quadrant represents teacher model and input pair region that are both informative and trackable, indicating that despite existence of moderate correlations, approximate trends can still be captured. This suggests that while our framework may not be directly applicable in all practical settings, it remains useful in scenarios where correlations are numerically weak, allowing for approximate tracking of dynamic behavior.

As an example, Fig.~\ref{fig:dynamics_mnist} presents a trajectory for an informative model that falls in the first quadrant. Even beyond our evaluation training step of, $\tau=1000$, the observed dynamics provide feasible results.

\section{Conclusion}\label{Conclusion}
Previous studies have examined the dynamics of neural networks under simple Gaussian distributions and Gaussian mixtures without hidden manifolds \citep{goldt2020modeling, goldt2022gaussian, pmlr-v139-refinetti21b_plain, loureiro2021learning_batch, univer4}. These works identified a form of Gaussian universality that facilitates analytical approaches to deep learning dynamics.

In this work, we extended the analysis by incorporating more general Gaussian mixtures into the hidden manifold framework and examined their impact. By comparing these dynamics to those under simple Gaussian inputs, we highlighted the role of input distribution complexity and its influence on convergence behavior.\\

Our central finding, strongly supported by numerical evidence, is that the Gaussian equivalence property exhibits remarkable robustness. It persists under more general input structures, including block-dependent Gaussian mixtures, provided that the data are standardized and exhibit weak correlations. Standardization aligns up to second--order cumulants with those of a Gaussian distribution, while the weak correlation condition allows internal random variables statistics to approximate its Gaussian counterparts. This convergence reveals a broadened form of universality in neural network dynamics, grounded in low-order statistical structure rather than exact distributional form.

Additionally, numerical experiments using a real dataset, where the teacher model input is structured rather than purely random, provided empirical insights into the role of correlation in convergence. By analyzing various teacher model and input pairs, we demonstrated the potential applicability of our framework in capturing real-world scenarios to some extent.

These results suggest that the conditions ensuring Gaussian universality in our framework may be looser than previously assumed, potentially holding even under weaker assumptions.

Future directions include applying this framework to more diverse datasets and relaxing constraints such as boundedness or specific correlation structures.

Another avenue is to examine effects of high order statistics. Specifically, our study suggests that even in the presence of higher-order cumulants, such as those introduced by Gaussian mixture components, learning dynamics can remain approximately invariant, provided these cumulants are attenuated in the thermodynamic or weakly correlated limit. This points to a potentially fundamental role for low-order cumulants in determining the effective behavior of deep learning models.

Recent studies have also begun to explore the impact of higher-order statistics to the localized breakdown of Gaussian equivalence \citep{hocs1, hocs2}. Together with our findings, this raises the open question of how far such equivalence principles can be extended in the presence of non-Gaussianity, especially in regimes where higher-order cumulants play a significant role.

From a different perspective, the conditional equivalence principle is closely connected to Gaussian mixture frameworks, a connection explored in \citep{nonG1} based on concepts from \citep{superstat}. In particular, the equivalent feature map in such models resembles an infinite-component mixture indexed by spike projections. This connection implies that our findings may also be relevant in the context of spiked models and other forms of input heterogeneity.

Exploring how deep learning parameters behave can be key to understanding the generalization ability of deep learning. We expect that our research will contribute to bridging the gap between the practical successes of applied deep learning and its developing theoretical foundations.

\section{Reproducibility}

For a comprehensive understanding of our numerical SGD implementation and ODE update mechanisms and to ensure reproducibility, please visit our code repository at \url{https://github.com/peardragon/GaussianUniversality}.

\begin{acknowledgments}
This work was supported by the Ministry of Education of the Republic of Korea and the National Research Foundation of Korea(NRF-2018S1A3A2075175)
\end{acknowledgments}

\appendix

\setcounter{figure}{0}
\setcounter{table}{0}
\setcounter{section}{0}
\setcounter{equation}{0}


\renewcommand\thefigure{A\arabic{figure}}
\makeatletter

\section{Derivation of Gaussian equivalence property and ODE}\label{section:background}
This appendix section rephrases the research from \citet{goldt2020modeling} to align with the notation used in the main script. For match the subscripts of manuscripts, original super(sub)scripts are reorganized.
\subsection{Correlation of Two Functions}
It is important to consider how to express the correlation of functions, such as $\mathbb{E}[f(x)g(y)]$, for the analysis of neural network dynamics.
Let's consider random variables following a $\mathcal{N}(0,1)$ distribution and examine the correlation of functions taking these random variables as inputs.

Represent two random variables, adhering to a joint Gaussian distribution, as vectors,
\begin{equation}
    x = (x_1, \cdots,x_I)^\top, \quad y = (y_1, \cdots, y_J)^\top.
\end{equation}
The assumption of joint Gaussian distribution for these random variables implies that the vectors have the following mean and covariance.
\begin{equation}
    \begin{aligned}
    \mathbb{E}[x_i]&=\mathbb{E}[y_j]=0,\\
    \mathbb{E}[x_ix_j]&= Q_{ij}, \quad \mathbb{E}[y_iy_j]=R_{ij}, \quad \mathbb{E}[x_iy_j] = \epsilon S_{ij}
\end{aligned}
\end{equation}
The joint distribution of $x$ and $y$ can be represented as:
\begin{equation}
P=\frac{1}{Z} \exp \left[-\frac{1}{2}\left(\begin{array}{ll}
x & y
\end{array}\right)\left(\begin{array}{cc}
Q & \epsilon S \\
\epsilon S^{\top} & R
\end{array}\right)^{-1}\left(\begin{array}{l}
x \\
y
\end{array}\right)\right]
\end{equation}
Considering a first-order approximation in $\epsilon$, the inverse matrix part becomes,
\begin{equation}
    \left(\begin{array}{ll}
Q & \epsilon S \\
\epsilon S^\top & R
\end{array}\right)^{-1} = \left(\begin{array}{ll}
Q &  0 \\
 0 & R
\end{array}\right)^{-1} - \epsilon M + \mathcal{O}(\epsilon^2)
\end{equation}
where \begin{equation}
    M = \left(\begin{array}{ll}
0 & Q^{-1} S R^{-1} \\
\left[Q^{-1} S R^{-1}\right]^\top & 0
\end{array}\right).
\end{equation}
Inserting this back into the joint distribution and approximating again with respect to $\epsilon$, we obtain following results.
\begin{equation}
    P(x, y)=\frac{1}{Z} \exp \left[A + B\right]
\end{equation}
where
\begin{equation}
    A = -\frac{1}{2}\left(\begin{array}{ll}x & y\end{array}\right)\left(\begin{array}{cc}Q^{-1} & 0 \\0 & R^{-1}\end{array}\right)\left(\begin{array}{l}x \\y\end{array}\right)
\end{equation}
and
\begin{equation}
    B = 1+\varepsilon \sum_{i=1}^I \sum_{j=1}^J x_i\left(Q^{-1} S R^{-1}\right)_{i j} y_j+\mathcal{O}\left(\varepsilon^2\right).
\end{equation}
To directly apply the aforementioned equation to the correlation of two functions, consider $f(x)$ and $g(y)$ as functions of $x$ and $y$, respectively. Provided these functions are sufficiently regular to possess expectations $\mathbb{E}_x[x_if(x)]$, $\mathbb{E}y[y_jg(y)]$, $\mathbb{E}x[x_ix_jf(x)]$, and $\mathbb{E}y[y_iy_jg(y)]$, the correlation between the two functions $\mathbb{E}[f(x)g(y)]$ can be expressed as:
\begin{widetext}
\begin{equation}\label{appendix:functionapprox}
    \mathbb{E}[f(x)g(y)] =  \mathbb{E}[f(x)]\mathbb{E}[g(y)]+ \epsilon \sum_{i=1}^I \sum_{j=1}^J \mathbb{E}[x_if(x)] (Q^{-1}SR^{-1})_{ij}\mathbb{E}[y_jg(y)] + \mathcal{O}(\epsilon^2),
\end{equation}
\end{widetext}
as derived by \citep{goldt2020modeling}.

\subsection{Gaussian Equivalence Property}
From the function correlation approximations, it becomes clear that for functions of sufficient regularity, their correlations are primarily dictated by the function's mean, distribution characteristics such as $\mathbb{E}[uf(u)]$, and the covariance of the original random variables. This underscores the pivotal role of function correlation in dissecting the dynamics within neural networks. We summarized previous results derived by \citet{goldt2020modeling}.

In our investigation, the weight update mechanism is facilitated by employing a straightforward stochastic gradient descent (SGD) strategy, with the batch size set to one.
\begin{align}
W_{k,i} & :=W_{k,i}-\frac{\eta}{\sqrt{N}} v_k (\hat{y}-y) g^{\prime}\left(\lambda_k\right) f(U_i) \\v_k &:= v_k -\frac{\eta}{N} g\left(\lambda_k\right) (\hat{y}-y)
\end{align}

By defining the normalized number of steps as $t = 1/N$ within the thermodynamic limit as $N \to \infty$, which analogously functions as a continuous time-like variable, we are equipped to elucidate the dynamics of the second layer weight in the student model by examining the function correlations of the preactivations from an averaged standpoint. Consequently, the dynamics of $v_k$ adhere to the following ODE formulation.
\begin{equation}
    \frac{dv_k}{dt} = \eta\left[ \sum^M_n \widetilde v_n I_1 -\sum^K_j  v_j I_2 \right ]
\end{equation}
where
\begin{equation}
\begin{aligned}
    I_1 = \mathbb{E}[g(\lambda_k) \widetilde{g}(\nu_n)], \quad 
    I_2 = \mathbb{E}[g(\lambda_k){g}(\lambda_j) ].
\end{aligned}
\end{equation}
Given the crucial role of function correlation in unpacking the dynamics prompted by weight updates, it is imperative to understand the distribution characterizing ${\lambda, \nu}$ to compute expectation values such as $\mathbb{E}[g(\lambda_k)\widetilde{g}(\nu_n)]$. This analytical approach enables a deeper understanding of the underlying mechanics governing the behavior of neural networks, particularly in how weight adjustments influence overall learning and adaptation processes.

Unlike the earlier discussion on simple function correlation, where the variable $x$ of the function was assumed to be a simple Gaussian, in the context of deep learning SGD updates, the random variable entering the function is not just an assumable random variable but the preactivations.

Therefore, it’s essential to ascertain the distribution of these preactivations. We need to starting following assumptions, Assumption~\ref{assumption:goldt:F} and Assumption~\ref{assumption:goldt:bounded}, taken from \citet{goldt2020modeling}.

In typical deep learning scenarios, activations that address gradient vanishing or explosiveness involve gradients directly influencing weight updates in a non-vanishing limit. Thus, considering bounds for student weights during initialization is sufficient. Since the remaining teacher weights and feature matrix $F$ are constant, ensuring proper bounds for teacher and student weights during initialization, and setting the feature matrix $F$ to be sufficiently bounded, these assumptions can be adequately met.

With these assumptions and the result of function correlation, the Gaussian Equivalence Property holds as follows:
\begin{property}[Gaussian Equivalence Property (GEP)]\label{appendix:GEP}
In the thermodynamic limit ($N\to \infty$, $D \to \infty$), with finite $K$, $M$, $D/N$, and under the Assumption~\ref{assumption:goldt:bounded} and Assumption~\ref{assumption:goldt:F}, if the $C$ follows a normal Gaussian distribution $\mathcal{N}(0,I)$, then $\{\lambda, \nu \}$ conform to $K+M$ jointly Gaussian variables. This means that statistics involving $\{\lambda, \nu\}$ are entirely represented by their mean and covariance.
\end{property}
This property derived by \citet{goldt2020modeling} allows us to representing characteristics of the student and teacher models, generalization error and dynamics of the student model's second layer weights, through the mean and covariance of the joint Gaussian distribution of preactivations.

For convenience, let's redefine $\overline{\lambda}_k$ as:
\begin{equation}
    \overline{\lambda}_k = \cfrac{1}{\sqrt{N}} \sum_{i=1}^N W_{k,i} ( f(U_i)-\mathbb{E}_{u \sim \mathcal{N}(0,1)}[f(u)])
\end{equation}
$\overline{\lambda}_k$ also follows a jointly Gaussian distribution, and its expectation value satisfies $\mathbb{E}[\overline{\lambda}_k]=0$ as per function correlation.

In this appendix, we present a concise derivation of $Q_{k, \ell}$. For a additional derivation, we refer the reader to prior research \citep{goldt2019dynamics}.
To facilitate the explanation, we first define $a$, $b$, and $c$ as statistical properties of the nonlinear function $f$, which is utilized in transforming the student model inputs $X$, where $X=f(CF/\sqrt{D})$:
\begin{equation}
    \begin{aligned}
    a &= \mathbb{E}_{u \sim \mathcal{N}(0,1)}[f(u)],\\
    b &= \mathbb{E}_{u \sim \mathcal{N}(0,1)}[uf(u)],\\
    c &= \mathbb{E}_{u \sim \mathcal{N}(0,1)}[f(u)^2]
    \end{aligned}
\end{equation}

With these definitions in place, $Q_{k, \ell}$ can be expressed as follows:

\begin{equation}
    \begin{aligned}
    Q_{k, \ell} &\equiv \mathbb{E}\left[{\overline\lambda}_k {\overline\lambda}_{\ell}\right] \\
    &= \mathbb{E}[\frac{1}{N}\sum^N_{i=1} \sum^N_{j=1} W_{k,i} W_{\ell, j} (f(U_i-a)(f(U_j)-a))]
    \end{aligned}
\end{equation}

Considering the case where $i \neq j$, and applying the expectation, we implement the function correlation approximation \ref{appendix:functionapprox} to derive:
\begin{equation}
    \mathbb{E}[f(U_i)f(U_j)] = a^2 + \frac{1}{D}\sum^D_{r=1} F_{r,i}F_{r,j}b^2
\end{equation}
in thermodynamic limit or
\begin{equation}
    \mathbb{E}[(f(U_i)-a)(f(U_j)-a)] = \frac{1}{D}\sum^D_{r=1} F_{r,i}F_{r,j}b^2.
\end{equation}

Hence, $Q_{k, \ell}$ can be succinctly rearranged for both $i \neq j$ and $i = j$ cases as:
\begin{widetext}\label{appendix:Q}
    \begin{align}
        Q_{k, \ell} &= (c-a^2)\frac{1}{N}\sum^N_{i=j=1}W_{k,i} W_{\ell, j} + \frac{1}{N}\sum^N_{i \neq j} W_{k,i} W_{\ell, j} [b^2\frac{1}{D}\sum^D_{r=1} F_{r,i}F_{r,j} ] \\
        & = (c-a^2-b^2)\frac{1}{N}\sum^N_{i=j=1}W_{k,i} W_{\ell, j} + \frac{1}{N}\sum^N_{i,j} W_{k,i} W_{\ell, j} [b^2\frac{1}{D}\sum^D_{r=1} F_{r,i}F_{r,j} ].
    \end{align}    
\end{widetext}

A similar approach can be applied to derive the remaining covariance components. Regarding high-order moments, an analogous method is employed by extending the function correlation approximation results, Eq.~(\ref{appendix:functionapprox}) to more general cases, thereby demonstrating that such preactivations follow a Gaussian distribution in the thermodynamic limit. For a comprehensive explanation of this process, the reader is encouraged to consult the referenced research \citep{goldt2020modeling}.

Consequently, the new distribution $\{\overline{\lambda}, \nu\}$ follows a more straightforward distribution with the mean
\begin{equation}
    \mathbb{E}\left[\overline{\lambda}_k\right]=\mathbb{E}\left[\nu_m\right]=0
\end{equation}
and the covariance
\begin{align}
Q_{k, \ell} &\equiv \mathbb{E}\left[{\overline\lambda}_k {\overline\lambda}_{\ell}\right]=\left(c-a^2-b^2\right) \Omega_{k, \ell}+b^2 \Sigma_{k, \ell} \\R_{k, m} &\equiv \mathbb{E}\left[\overline{\lambda}_k \nu_m\right]=b \frac{1}{D} \sum_{r=1}^D S_{k,r} \widetilde{W}_{m,r} \\T_{m, n}& \equiv \mathbb{E}\left[\nu_m \nu_n\right]=\frac{1}{D} \sum_{r=1}^D \widetilde{W}_{m,r} \widetilde{W}_{n,r}.
\end{align}

The newly defined matrices satisfy the following relations:
\begin{align}
S_{k,r} & \equiv \frac{1}{\sqrt{N}} \sum_{i=1}^N W_{k,i} F_{r,i}, 
\\\Omega_{k, \ell} & \equiv \frac{1}{N} \sum_{i=1}^N W_{k,i} W_{\ell,i}, 
\\\Sigma_{k, \ell} & \equiv \frac{1}{D} \sum_{r=1}^D S_{k,r} S_{\ell, r}.
\end{align}

\subsection{Derivation of the ODE}
To derive the ODE for our main metrics of interest - the covariances $Q$, $R$, and the second layer weight $v$ - we begin with our single batch gradient update. 
\begin{align}
    W_{k,i} & :=W_{k,i}-\frac{\eta}{\sqrt{N}} v_k (\hat{y}-y) g^{\prime}\left(\lambda_k\right) f(U_i), \\v_k &:= v_k -\frac{\eta}{N} g\left(\lambda_k\right) (\hat{y}-y)
\end{align}

The preactivations are related to the first layer weights, and thus we consider quantities such as $S_{k,r}$ and $\Sigma_{k,\ell}$ that are proportional to the first layer weights $W$. The dynamics of the first layer weights are determined by a term involving $(\hat{y}-y) g'(\lambda_k)f(U_i)$, assuming the second layer is constant. The average update of these quantities can be obtained from the following equation:
\begin{equation}
    \left[\sum_{j=1}^Kv_j g(\lambda_j) - \sum_{m=1}^M \widetilde{v}_m \widetilde{g}(v_m) \right]g'(\lambda_k)f(U_i)
\end{equation}
Starting with $S_{k,r} \equiv \frac{1}{\sqrt{N}} \sum_{i=1}^N W_{k,i} F_{r,i}$, we obtain:
\begin{widetext}
    \begin{equation}
    S_{k,r}:=S_{k,r}-\cfrac{\eta}{\sqrt{N}}v^k\biggl[ \sum_{j\neq k}^K v_j \mathbb{E}[g(\lambda_j)g'(\lambda_k)\beta_r] + v_k\mathbb{E}[g(\lambda_k)g'(\lambda_k)\beta_r] -\sum_n^M\widetilde{v}_n\mathbb{E}[\widetilde{g}(\nu_n)g'(\lambda_k)\beta_r]\biggr]
    \end{equation}
\end{widetext}
with $\beta_r = \cfrac{1}{\sqrt{N}}\sum_i F_{r,i}f(U_i)$.

Function correlations are employed to express these updates in terms of statistical quantities of the distributions ${\lambda, \nu }$. However, the equations for covariances remain coupled. To uncouple them, we need to consider the eigenvectors and eigenvalues, $\psi_\tau$ and $\rho_\tau$, of the $D \times D$ matrix $\mathcal{F}$ formed by $\mathcal{F}_{r,s} ={1}/{N} \sum_i F_{r,i} F_{s,i}$. The eigenvectors and eigenvalues are obtained under the following normalization condition:
\begin{equation}
\begin{aligned}
    \sum_s \mathcal{F}_{r,s}(\psi_\tau)_s = \rho_\tau (\psi_\tau)_r, \\ \sum_s (\psi_\tau)_s (\psi_{\tau'})_s = D\delta(\tau,\tau'), \\ \sum_\tau (\psi_\tau)_r (\psi_{\tau})_s = D\delta(r,s)
\end{aligned}
\end{equation}\label{append:decoupling}
Using these, we can express the teacher-student overlap covariance $R_{k,m}$ through two projected matrices:
\begin{equation}
    \mathcal{S}_{k,r} = \cfrac{1}{\sqrt{D}}\sum_r S_{k,r} (\psi_\tau)_r, \quad \mathcal{W}_{m, \tau} = \cfrac{1}{\sqrt{D}}\sum_r \widetilde{W}_{m, r} (\psi_\tau)_r
\end{equation}
and thus:
\begin{equation}
    R_{k,m} =\cfrac{b}{D}\sum_\tau \mathcal{S}_{k,r} \mathcal{W}_{m, \tau}
\end{equation}
Since the teacher model's matrix is static, its projection matrix $\mathcal{S}$ is given by:
\begin{widetext}
\begin{equation}
        \mathcal{S}_{k, \tau} := \mathcal{S}_{k, \tau}-\cfrac{\eta}{\sqrt{DN}} v^k\sum_r (\psi_\tau)_r  \biggl[ \sum_{j\neq k}^K v_j \mathbb{E}[g(\lambda_j)g'(\lambda_k)\beta_r] + v_k\mathbb{E}[g(\lambda_k)g'(\lambda_k)\beta_r]  -\sum_n^M\widetilde{v}_n\mathbb{E}[\widetilde{g}(\nu_n)g'(\lambda_k)\beta_r]\biggr]
\end{equation}  
\end{widetext}

The update rule for $R$ is then derived using these projections. Explicitly at timestep $t$, it can be expressed as:
\begin{equation}
    (R_{k,m})_{t+1}-(R_{k,m})_{t} = \cfrac{b}{D}\sum_\tau \left[ (\mathcal{S}_{k, \tau})_{t+1}- (\mathcal{S}_{k, \tau})_{t}\right]\widetilde{W}_{m, r}
\end{equation}
as derived by \citet{goldt2020modeling}.
During the summation over $\tau$, two types of terms emerge:
\begin{equation}
    \mathcal{T}_{m,n} \equiv \cfrac{1}{D} \sum_\tau \rho_\tau \widetilde{W}_{m,r}\widetilde{W}_{n,r}, \quad \cfrac{1}{D}\sum_\tau \rho_\tau \mathcal{S}_{\ell, \tau}\widetilde{W}_{n, \tau}
\end{equation}
The second summation is not readily reducible to a simpler expression. Instead, we introduce the following density function:
\begin{equation}
    r_{k,m}(\rho) = \cfrac{1}{\epsilon_\rho}\cfrac{1}{D} \sum_\tau \widetilde{S}_{k,\tau} \widetilde{W}_{m, \tau} \mathbf{1}_{\rho_\tau \in [\rho, \rho+\epsilon_\rho]}
\end{equation}
This density function allows us to express the covariance $R$ in terms of the eigenvalue distribution $\rho$:
\begin{equation}
    R_{k,m} = b \int d\rho p(\rho)r_{k,m}(\rho)
\end{equation}
Under the assumption that the feature matrix elements are i.i.d. from a normal distribution $\mathcal{N}(0,1)$, this distribution adheres to the Marchenko-Pastur law \citep{marchenko1967distribution}:
\begin{equation}
    p(\rho) = \cfrac{\sqrt{((1+\sqrt{D/N})^2-\rho)(\rho-(1-\sqrt{D/N})^2)}}{2\pi D/N\, \rho}.
\end{equation}
The update equation for $r_{k,m}(\rho)$ is straightforwardly derived from the update equation and definition of $\mathcal{S}$. Ultimately, in the thermodynamic limit, with $t=1/N$ transforming into a continuous time-like variable, the equation of motion for $r_{k,m}(\rho, t)$ satisfies the following ODE:
\begin{widetext}
    \begin{equation}
    \begin{aligned}
    \frac{\partial r_{k, m}(\rho, t)}{\partial t}
    &=-\frac{\eta}{D/N} v_k d(\rho)\Biggl(  r_{k m}(\rho) \sum_{j \neq k}^K v_j \frac{Q_{j j} \mathbb{E}[g'(\lambda_k)\lambda_k g(\lambda_j)]-Q_{k j} \mathbb{E}[g'(\lambda_k)\lambda_j g(\lambda_j)]}{Q_{j j} Q_{k k}-\left(Q_{k j}\right)^2}
    \\&+\sum_{j \neq k}^K v_j r_{j m}(\rho) \frac{Q_{k k} \mathbb{E}[g'(\lambda_k)\lambda_j g(\lambda_j)]-Q_{k j} \mathbb{E}[g'(\lambda_k)\lambda_k g(\lambda_j)]}{Q_{j j} Q_{k k}-\left(Q_{k j}\right)^2} 
    \\& +v_k r_{k m}(\rho) \frac{1}{Q_{k k}} \mathbb{E}[g'(\lambda_k)\lambda_k g(\lambda_k)]-
    \\&r_{k m}(\rho) \sum_n^M \widetilde{v}_n \frac{T_{n n} \mathbb{E}[g'(\lambda_k)\lambda_k \widetilde{g}(\nu_n)]-R_{k n} \mathbb{E}[g'(\lambda_k)\nu_n \widetilde{g}(\nu_n)]}{Q_{k k} T_{n n}-\left(R_{k n}\right)^2} 
    \\&-\frac{b \rho}{d(\rho)} \sum_n^M \widetilde{v}_n \mathcal{T}_{n m} \frac{Q_{k k}\mathbb{E}[g'(\lambda_k)\nu_n \widetilde{g}(\nu_n)]-R_{k n} \mathbb{E}[g'(\lambda_k)\lambda_k \widetilde{g}(\nu_n)]}{Q_{k k} T_{n n}-\left(R_{k n}\right)^2}\Biggr)
    \end{aligned}
\end{equation}
\end{widetext}

where $d(\rho) = (c-b^2)\cfrac{D}{N} + b^2 \rho$ as derived by \citet{goldt2020modeling}.
Note that all explicit time dependencies on the right side of the equation are omitted for clarity. In this numerical ODE implementation, the right side corresponds to the immediate preceding time $t$, and the left side to the updated time $t+1$.

Similarly, the covariance $Q$ associated with the first weight $W$ can be derived in a repetitive manner, starting from:
\begin{equation}
    Q_{k,\ell} \equiv \mathbb{E}[\lambda_k\lambda_\ell] = [c-b^2]W_{k,\ell} + b^2 \Sigma_{k, \ell}
\end{equation}
Notice that we ignore $a$ term since we focus on symmetric nonlinear function $f$.
Following a similar process as before, we find that the first term, $W_{k,l}$, adheres to:
\begin{widetext}
\begin{equation}
    \begin{aligned}
    \frac{\mathrm{d} W_{k, \ell}(t)}{\mathrm{d} t}= & -\eta v_k\left(\sum_j^K v_j \mathbb{E}[g'(\lambda_k)\lambda_\ell {g}(\lambda_j)]-\sum_n \widetilde{v}_n \mathbb{E}[g'(\lambda_k)\lambda_\ell \widetilde{g}(\nu_n)]\right)
    \\&-\eta v_{\ell}\left(\sum_j^K v_j \mathbb{E}[g'(\lambda_\ell)\lambda_k {g}(\lambda_j)]-\sum_n \widetilde{v}_n \mathbb{E}[g'(\lambda_\ell)\lambda_k \widetilde{g}(\nu_n)]\right) 
    \\& +c \eta^2 v_k v_{\ell}\biggl(\sum_{j, \iota}^K v_j v_\iota \mathbb{E}[g'(\lambda_k)g'(\lambda_\ell)g(\lambda_j){g}(\lambda_\iota)]
    \\&-2 \sum_j^K \sum_m^M v_j \widetilde{v}_m \mathbb{E}[g'(\lambda_k)g'(\lambda_\ell)g(\lambda_j)\widetilde{g}(\nu_m)]
    \\&+\sum_{n, m}^M \widetilde{v}_n \widetilde{v}_m \mathbb{E}[g'(\lambda_k)g'(\lambda_\ell)\widetilde{g}(\nu_n)\widetilde{g}(\nu_m)]\biggr)
    \end{aligned}
\end{equation}
\end{widetext}
as derived by \citet{goldt2020modeling}.
The second term, $\Sigma_{k,\ell}$, can be expressed using the rotating basis $\psi_\tau$:
\begin{equation}
    \Sigma_{k,\ell} \equiv \cfrac{1}{D} \sum_r S_{k,r} S_{\ell,r} = \cfrac{1}{D} \sum_\tau \mathcal{S}_{k, \tau} \mathcal{S}_{\ell, \tau}
\end{equation}
and thus, integral form for $\Sigma_{k, \ell}(t)$ can be derived:
\begin{equation}
    \sigma_{k,\ell}(\rho) = \cfrac{1}{\epsilon_\rho}\cfrac{1}{D} \sum_\tau \mathcal{S}_{k,\tau} \mathcal{S}_{\ell, \tau} \mathbf{1}_{\rho_\tau \in [\rho, \rho+\epsilon_\rho]}
\end{equation}
with
\begin{widetext}
    \begin{equation}
    \begin{aligned}
    & \frac{\partial \sigma_{k \ell}(\rho, t)}{\partial t}=-\frac{\eta}{D/N}\left(d(\rho) v_k \sigma_{k \ell}(\rho) \sum_{j \neq k} v_j \frac{Q_{j j} \mathbb{E}[g'(\lambda_k)\lambda_k {g}(\lambda_j)]-Q_{k j} \mathbb{E}[g'(\lambda_k)\lambda_j {g}(\lambda_j)]}{Q_{j j} Q_{k k}-\left(Q_{k j}\right)^2}\right. 
    \\&+v_k \sum_{j \neq k} v_j d(\rho) \sigma_{j \ell}(\rho) \frac{Q_{k k} \mathbb{E}[g'(\lambda_k)\lambda_j {g}(\lambda_j)]-Q_{k j} \mathbb{E}[g'(\lambda_k)\lambda_k {g}(\lambda_j)]}{Q_{j j} Q_{k k}-\left(Q_{k j}\right)^2} 
    \\&+d(\rho) v_k \sigma_{k \ell}(\rho) v_k \frac{1}{Q_{k k}} \mathbb{E}[g'(\lambda_k)\lambda_k {g}(\lambda_k)] 
    \\&-d(\rho) v_k \sigma_{k \ell}(\rho) \sum_n \widetilde{v}_n \frac{T_{n n} \mathbb{E}[g'(\lambda_k)\lambda_k \widetilde{g}(\nu_n)]-R_{k n} \mathbb{E}[g'(\lambda_k)\nu_n \widetilde{g}(\nu_n)]}{Q_{k k} T_{n n}-\left(R_{k n}\right)^2} 
    \\&-b \rho v_k \sum_n \widetilde{v}_n r_{\ell n}(\rho) \frac{Q_{k k} \mathbb{E}[g'(\lambda_k)\nu_n \widetilde{g}(\nu_n)]-R_{k n} \mathbb{E}[g'(\lambda_k)\lambda_k \widetilde{g}(\nu_n)]}{Q_{k k} T_{n n}-\left(R_{k n}\right)^2} 
    \\&+ \text { all of the above with } \ell \rightarrow k, k \rightarrow \ell) . 
    \\&+\eta^2 v_k v_{\ell}\left[\left(c-b^2\right) \rho+\frac{b^2}{\delta} \rho^2\right]\left(\sum_{j, \iota}^K v_j v_\iota \mathbb{E}[g'(\lambda_k)g'(\lambda_\ell)g(\lambda_j){g}(\lambda_\iota)]\right. 
    \\&\left.-2 \sum_j^K \sum_m^M v_j \widetilde{v}_m \mathbb{E}[g'(\lambda_k)g'(\lambda_\ell)g(\lambda_j)\widetilde{g}(\nu_m)+\sum_{n, m}^M \widetilde{v}_n \widetilde{v}_m \mathbb{E}[g'(\lambda_k)g'(\lambda_\ell)\widetilde{g}(\nu_n)\widetilde{g}(\nu_m)]\right)
    \end{aligned}
\end{equation}
\end{widetext}
as derived by \citet{goldt2020modeling}.
The weight $v$ and generalization error $\epsilon_g$ can be directly obtained from the weight update formula and the definition of generalization error with MSE:
\begin{equation}
    \frac{dv_k}{dt} = \eta\left[ \sum^M_n \widetilde v_n \mathbb{E}[g(\lambda_k) \widetilde{g}(\nu_n)] -\sum^K_j  v_j \mathbb{E}[g(\lambda_k){g}(\lambda_j) ] \right ]
\end{equation}
with
\begin{widetext}
\begin{equation}
\begin{aligned}
        \epsilon_g(\theta, \widetilde{\theta})
        &=\frac{1}{2} \mathbb{E}\left[ \left(\sum_k^K v_k g\left(\lambda_k\right)-\sum_m^M \widetilde{v}_m \widetilde{g}\left(\nu_m\right)\right)^2\right]
        \\&=\frac{1}{2} \sum_{k, \ell} v_k v_{\ell} \mathbb{E}[g(\lambda_k) {g}(\lambda_\ell)]+\frac{1}{2} \sum_{n, m} \widetilde{v}^n \widetilde{v}^m \mathbb{E}[\widetilde{g}(\nu_n) \widetilde{g}(\nu_m)]-\sum_{k, n} v_k \widetilde{v}_n \mathbb{E}[g(\lambda_k) \widetilde{g}(\nu_n)]
\end{aligned}
\end{equation}
\end{widetext}

\section{Function Correlation Decomposition}\label{Appendix:FunctionDecompose}
\subsection{Vector and Block Structure}\label{blockstructure}
Let $D$ be the ambient dimension and partition indices $\{1,\dots,D\}$ into $n_b$ (mutually) independent blocks
\begin{equation}
\{\mathcal B_1,\dots,\mathcal B_{n_b}\},\qquad |\mathcal B_b|\le m,\qquad n_b \asymp D/m.
\end{equation}
Let $\CbarBm=(\cbarBm_1,\dots,\cbarBm_D)^\top$ be a global standardized random vector with
\begin{equation}
\E\,\cbarBm_r=0,\quad \Var(\cbarBm_r)=1,
\end{equation}
and block wise covariance matrix
\begin{equation}
\Sigma_b:=\Cov(\CbarBm_{\mathcal B_b})
\end{equation}
satisfying the uniform spectral bounds,
\begin{equation}\label{eq:block-spectral-bds}
  c_\Sigma I \preceq \Sigma_b \preceq C_\Sigma I \qquad \text{for all } b,
\end{equation}
for constants $0<c_\Sigma\le C_\Sigma<\infty$.
Given deterministic feature matrix $F\in\R^{D\times N}$, under assumption
\begin{assumption}[Bound assumption of $F$]\label{eq:sum-form-coherence}
\begin{equation*}
  \sum_{r=1}^D F_{r,i}^2 \;=\; D,\quad
  \frac{1}{\sqrt D}\sum_{r=1}^D F_{r,i}F_{r,j} \;=\; \bigO(1)\quad \text{for } i\neq j,
\end{equation*}
\end{assumption}
\subsection{Exchangable Pair Definition}
Let $S\in\R^p$ be any statistic that decomposes as a sum of independent, mean-zero block contributions,
\begin{equation}\label{eq:S-sum}
  S \;=\; \sum_{b=1}^{n_b} W_b,
  \quad
  W_b \;=\; \Phi_b\!\big(\CbarBm_{\mathcal B_b}\big)\;-\;\E\Big[\Phi_b\!\big(\CbarBm_{\mathcal B_b}\big)\Big]
  ,
\end{equation}
where each $\Phi_b:\R^{|\mathcal B_b|}\to\R^p$ is a map depending only on the block coordinates
$\CbarBm_{\mathcal B_b}$. By block independence, the vectors $\{W_b\}_{b=1}^{n_b}$ are independent, and by construction $\E W_b=0$.

Construct an exchangeable pair $(S,S')$ by block resampling, draw $I\sim\mathrm{Unif}\{1,\dots,n_b\}$, replace $\CbarBm_{\mathcal B_I}$ by an i.i.d.\ copy $\overline{\mathbf{C}}'^{\mathcal B}_{\mathcal B_I}$ to form $W'_I$, and set
\begin{equation}
S' := S - W_I + W'_I,\quad D := S'-S=W'_I-W_I.
\end{equation}
Then $(S,S')$ is exchangeable and satisfies the exact regression
\begin{equation}\label{eq:regression}
  \E[D\mid S] \;= -\Lambda S=\; -\frac{1}{n_b}\,S.
\end{equation}
Let $\Sigma:=\Cov(S)$ and define the matrix error
\begin{equation}\label{eq:E-matrix}
  E \;:=\; \frac12\,\E\!\big[(\Lambda^{-1}D)D^\top\mid S\big] - \Sigma,
  \qquad D:=S'-S.
\end{equation}
\subsection{Bound Property of Expectation terms}
We first bound the two key quantities $\E\,\|\Lambda^{-1}D\|^3$ and $\E\,\|E\|_{\HS}$.
\subsubsection{Cubic Term}
Since $\Lambda^{-1}=n_b I_p$ and $D=W'_I-W_I$,
\begin{align}
\E\|\Lambda^{-1}D\|^3
&= n_b^3\,\E\|W'_I-W_I\|^3\\
&= n_b^3\cdot \frac{1}{n_b}\sum_{b=1}^{n_b}\E\|W'_b-W_b\|^3 \notag\\
&\le n_b^3\cdot \frac{1}{n_b}\sum_{b=1}^{n_b} \E\big(\|W'_b\|+\|W_b\|\big)^3\\
&\le 8\,n_b^2 \cdot \frac{1}{n_b}\sum_{b=1}^{n_b}\E\|W_b\|^3, \label{eq:cubic-triangle}
\end{align}
using $(x+y)^3\le 4(x^3+y^3)$ and $\E\|W'_b\|^3=\E\|W_b\|^3$.

For the any $S$ built on a fixed $p$-tuple of indices, write
\begin{equation}
W_b \;=\; A_b\,\xi_b,\qquad 
\xi_b:=\CbarBm_{\B_b}\in\R^{|\B_b|}, 
\end{equation}
where the $r$-th column of \(A_b\) is \(D^{-1/2}\,[F_{r,i_1},\ldots,F_{r,i_p}]^\top\) for $r\in\B_b$.
Let \(\Sigma_b=\Cov(\xi_b)\). Then
\begin{equation}
\E\|W_b\|^2
= \E\,\mathrm{tr}\big(W_bW_b^\top\big)
= \mathrm{tr}\big(A_b\,\Sigma_b\,A_b^\top\big).
\end{equation}

Assume that
\begin{assumption}[Block bound assumption 1]\label{assumption:energy}
\begin{equation}
        \mathrm{tr}\big(A_b\,\Sigma_b\,A_b^\top\big) \leq (m/D)^\beta.
\end{equation}
\end{assumption}

Then the second moment be bounded by
\begin{equation}
\E\|W_b\|^2 \;\leq (m/D)^\beta.
\end{equation}
Also, by definition fourth moment be bounded by
\begin{equation}
    \begin{aligned}
    \E\|W_b\|^4
= \E\big(\xi_b^\top B\,\xi_b\big)^2
&= 2\,\mathrm{tr}\!\big[(B\Sigma_b)^2\big] + \big(\mathrm{tr}(B\Sigma_b)\big)^2
\\ &\le\ 3\,\big(\mathrm{tr}(B\Sigma_b)\big)^2.
    \end{aligned}
\end{equation}

Using $\mathrm{tr}(B\Sigma_b)=\mathrm{tr}(A_b\Sigma_b A_b^\top)\leq(m/D)^{\beta}$ gives

\begin{equation}\label{eq:Wb-fourth}
  \E\|W_b\|^4 \;\leq (m/D)^{2\beta}.
\end{equation}

As results, third moment be bounded by
\begin{equation}\label{eq:Wb-cubic}
  \E\|W_b\|^3 \;\le\; \big(\E\|W_b\|^4\big)^{3/4} \leq (m/D)^{3\beta/2}.
\end{equation}

Finally,
\begin{equation}\label{eq:step-cubic}
    \begin{aligned}
\E\|\Lambda^{-1}D\|^3 &= \frac{1}{\lambda}\,\E\|D\|^3 \\
&=\; n_b\,\E\|D\|^3 \\
&\leq n_b  \frac{8}{n_b}\sum_{b=1}^{n_b}\E\|W_b\|^3\\
&\leq 8n_b \Bigl(\frac{m}{D}\Bigr)^{3\beta/2}
\end{aligned}
\end{equation}
where $n_b\asymp D/m$.
\subsubsection{Matrix Error Term}
Let $T:=\sum_{b=1}^{n_b}W_bW_b^\top$ so that $\E T=\Sigma$. A direct block computation yields
\begin{equation}
    \begin{aligned}
\E\!\big[(\Lambda^{-1}D)D^\top\mid S\big]
&= n_b\,\E\!\big[(W'_I-W_I)(W'_I-W_I)^\top\mid S\big]\\
&= \sum_{b=1}^{n_b}\Sigma_b \;+\; \sum_{b=1}^{n_b}\E[W_bW_b^\top\mid S],
    \end{aligned}
\end{equation}
hence, matrix error is expressed by
\begin{equation}\label{eq:E-T-diff}
  E \;=\; \frac12\big(\E[T\mid S]-\E T\big).
\end{equation}
Hilbert-Schmidt norm has inequality results to
\begin{equation}
\begin{aligned}
    \E\|E\|_{\HS}
\;&=\; \frac12\,\E\big\|\E[T\mid S]-\E T\big\|_{\HS}\\
\;&\le\; \frac12\Big(\E\big\|\E[T\mid S]-\E T\big\|_{\HS}^2\Big)^{1/2}.
\end{aligned}
\end{equation}
Using inequality $\E\big\|\E[Y\mid S]\big\|_{\HS}^2 \;\le\; \E\|Y\|_{\HS}^2$, 
\begin{equation}
\E\|E\|_{\HS} \;\le\tfrac12\sqrt{\Var(T)}.
\end{equation}

Using block wise independence,
\begin{equation}
    \begin{aligned}
 \Var(T)\;\defeq\;\sum_{i,j}\Var(T_{ij})
\;&\le\;\sum_{b}\sum_{i,j}\E\big[W_{b,i}^2 W_{b,j}^2\big]
\\&=\;\sum_{b}\E\Big[\Big(\sum_{i} W_{b,i}^2\Big)^2\Big]
\\&=\;\sum_{b}\E\big[\|W_b\|_2^4\big].
    \end{aligned}
\end{equation}
This also results to
\begin{equation}
    \Var(T)\ \le\ \sum_{b=1}^{n_b}\E\|W_b\|_2^{\,4}\ \leq n_b \Bigl(\frac{m}{D}\Bigr)^{2\beta},
\end{equation}
finally,
\begin{equation}\label{eq:E-HS}
  \E\|E\|_{\HS}
  \;\le\; \cfrac12\sqrt{\Var(T)}
  \;\leq \frac{1}{2}\sqrt{n_b} \Bigl(\frac{m}{D}\Bigr)^{\beta}
\end{equation}
\subsection{Approximation Expectation of Function}
In summary, we can use smooth function bound theorem
\begin{theorem}[Smooth function bound with improved dimension dependence]
Let $Z\sim\mathcal N(0,\Sigma)$. Random vector $S$ with zero mean and $\Cov(S) = \Sigma$. For any second order differentiable functions $h$ with bounded derivatives,
\begin{equation}\label{eq:Stein-master}
\begin{aligned}
    \big|\E h(S)-\E h(Z)\big|
\;\le\;
M_1(h)\,\|\Sigma^{-1/2}\|_{\op}\,\E\|E\|_{\HS} \\ 
\;+\; C_\star\,\|\Sigma^{-1/2}\|_{\op}\,M_3(h)\,\E\|\Lambda^{-1}D\|^3,
\end{aligned}
\end{equation}
where $M_k(h)=\sup_x\|\nabla^k h(x)\|_{\op}$ and $C_\star$ is an absolute constant (e.g.\ $1/(12\sqrt{2\pi})$).
By \eqref{eq:block-spectral-bds}, $\|\Sigma^{-1/2}\|_{\op}\le c_\Sigma^{-1/2}=\bigO(1)$. 
\end{theorem}

Substituting \eqref{eq:step-cubic} and \eqref{eq:E-HS} into \eqref{eq:Stein-master}, results to
\begin{equation}
\begin{aligned}
        \big|\E h(S)-\E h(Z)\big|
\;\le\;
&\underbrace{M_1(h)\,c_\Sigma^{-1/2}\,C_1}_{=:A_h}\,(\tfrac{m}{D})^{\beta - 1/2}
\;+\;\\
&\underbrace{C_\star\,c_\Sigma^{-1/2}\,M_3(h)\,C_2}_{=:B_h}\,(\tfrac{m}{D})^{3\beta/2 - 1}.
\end{aligned}
\end{equation}
Absorbing $A_h+B_h$ into $C_h$ and using $3\beta/2-1 = \gamma$ yields the final form.

\begin{lemma}[Smooth-test approximation]\label{lem:smooth-test}
For $S$ as in \eqref{eq:S-sum} and $Z\sim\mathcal N(0,\Sigma)$ with $\Sigma=\Cov(S)$, under \eqref{eq:block-spectral-bds} and \eqref{assumption:energy}, for any $h\in C^3(\R^p)$ with bounded derivatives,
\begin{equation}\label{eq:smooth-rate-detailed}
\big|\E h(S)-\E h(Z)\big|
\;\le\; C_h\left({\frac{m}{D}}\right)^\gamma
\end{equation}
\end{lemma}
\subsection{Applications}
\subsubsection{Linear Statistics}\label{application:Ui}
For example,
For a set of indices $\mathcal I=\{i_1,\dots,i_p\}\subset\{1,\dots,N\}$ and
\[
U_i \;=\; \frac{1}{\sqrt D}\sum_{r=1}^D F_{r,i}\,\overline{C}^{\mathcal{B}}_r,
\]
take $S=(U_{i_\ell})_{\ell=1}^p\in\R^p$ and define
\begin{equation}\label{eq:Wb-linear-example}
  W_b \;=\; \underbrace{\frac{1}{\sqrt D}\sum_{r\in\mathcal B_b} \overline{C}^{\mathcal{B}}_r\,
  \begin{bmatrix}
    F_{r,i_1}\\ \vdots\\ F_{r,i_p}
  \end{bmatrix}}_{\Phi_b\!\big(\CbarBm_{\mathcal B_b}\big)}
  \;-\; \E\!\left[\Phi_b\!\big(\CbarBm_{\mathcal B_b}\big)\right]
\end{equation}
so that $\E W_b=0$ (since $\E \overline{C}^{\mathcal{B}}_r=0$). Then $S=\sum_b W_b$ holds.
\subsubsection{Index Statistics}\label{application:Cr}
Let $b(r)$ be the block containing index $r$ and put
$\xi_b:=\CbarBm_{\mathcal B_b}$. Define for each block $b$ the $\R^2$-valued map
\begin{equation}
W_b \;:=\;
\begin{bmatrix}
w_b^{(c)}\\[2pt] w_b^{(u)}
\end{bmatrix} 
\end{equation}
with
\begin{equation}
w_b^{(c)}:=\mathbf 1_{\{b=b(r)\}}\big(\xi_b(\text{pos}(r))-\E\,\xi_b(\text{pos}(r))\big), 
\end{equation}
and
\begin{equation}
    w_b^{(u)}:=\frac{1}{\sqrt D}\sum_{s\in\mathcal B_b} F_{s,i}\,\xi_b(s),
\end{equation}
where $\text{pos}(r)$ is the local coordinate of $r$ inside $\mathcal B_{b(r)}$.
Then $\E W_b=0$ and, by construction,
\[
S\;:=\;\begin{bmatrix}\overline{C}^{\mathcal{B}}_r\\ U_i\end{bmatrix}
\;=\;\sum_{b=1}^{n_b} W_b.
\]

In summary, we can consider $S$ as joint distribution $(\overline{C}^{\mathcal{B}}_r, U_j)$ or $(U_i, U_j)$. This allows that we can approximate function correlations by the results from Gaussian with the same mean and covariance.

\subsection{Decomposition of Joint Gaussian Distribution Expectation}

Lets take 
random vector $Z_\rho = (X_\rho, Y_\rho)$ with mean and variance
$(X_\rho,Y_\rho)\sim\mathcal N\big((\mu_x,\mu_y), \Sigma(\rho))$ where off diagonal term has weak correlation,
\begin{equation}
\Sigma(\rho)=\begin{pmatrix}\Sigma_{xx} & \rho \Sigma_{xy}\\\rho \Sigma_{xy}^\top & \Sigma_{yy}\end{pmatrix}.
\end{equation}
Using Price's Theorem, 
\begin{equation}
    \frac{d}{d\rho}\,\mathbb E[\varphi(Z_\rho)]=\frac12\,\mathbb E\!\big[\nabla^2\varphi(Z_\rho):\Sigma'(\rho)\big],\quad Z_\rho=(X_\rho,Y_\rho), 
\end{equation}
with
\begin{equation}
    \Sigma'(\rho)=\begin{pmatrix}0&\Sigma_{xy}\\\Sigma_{yx}&0\end{pmatrix}.
\end{equation}
where
\begin{equation}
A:B \;:=\; \mathrm{tr}(A^\top B)\;=\;\sum_{i,j} A_{ij}B_{ij}.
\end{equation}
Choosing $\varphi(x,y)=f(x)g(y)$ yields
\begin{equation}\label{eq:price-first-deriv}
    h'(\rho)=\mathbb E\!\big[\nabla f(X_\rho)^\top\,\Sigma_{xy}\,\nabla g(Y_\rho)\big].
\end{equation}
For Gaussian $X_0\sim\mathcal{N}(\mu_x,\Sigma_{xx})$, $Y_0\sim\mathcal{N}(\mu_y,\Sigma_{yy})$, Stein's Lemma says that,
\begin{equation}\label{eq:stein-grad}
\begin{aligned}
     \mathbb E[\nabla f(X_0)]&=\Sigma_{xx}^{-1}\mathbb E[(X_0-\mu_x)f(X_0)],\\
     \mathbb E[\nabla g(Y_0)]&=\Sigma_{yy}^{-1}\mathbb E[(Y_0-\mu_y)g(Y_0)],   
\end{aligned}
\end{equation}
Plugging \eqref{eq:stein-grad} into Taylor expansion yields the compact form
\begin{widetext}
    \begin{equation}
    \begin{aligned}h(\rho)&=h(0)+\rho\,h'(0)+O(\rho^2)\\&=\mathbb E[f(X_0)]\mathbb E[g(Y_0)] +\rho\,\Big(\mathbb E[\nabla f(X_0)]^\top\Sigma_{xy}\,\mathbb E[\nabla g(Y_0)]\Big)+O(\rho^2)\\&=\mathbb E[f(X_0)]\mathbb E[g(Y_0)] +\rho\,\mathbb E[(X_0-\mu_x)f(X_0)]^\top\underbrace{\Sigma_{xx}^{-1}\Sigma_{xy}\Sigma_{yy}^{-1}}_{=:A}\, \mathbb E[(Y_0-\mu_y)g(Y_0)] +O(\rho^2).\end{aligned}
\end{equation}

From \ref{application:Ui}, \ref{application:Cr}, we can conclude this application  
\begin{equation}\label{euiuj}
\left|\mathbb{E}[f(U_i)f(U_j)] - \left( a^2 + \frac{1}{D}\sum^D_{r=1} F_{r,i}F_{r,j}b^2\right)\right| <  C\left({\frac{m}{D}}\right)^\gamma + \mathcal{O}(\rho^2)
\end{equation}
and
\begin{equation}\label{ecrui}
\left|\mathbb{E}[f(U_i)\cbarBm_r] - \left( \frac{1}{\sqrt{D}}F_{r,i} \mathbb{E}[U_i f(U_i)]\right)\right| <  C\left({\frac{m}{D}}\right)^\gamma + \mathcal{O}(\rho^2)
\end{equation}
where $\rho^2$ from weak correlation.
\end{widetext}

\section{Derivation of Gaussianity}\label{appendix:Gaussianity}
We work under the $m$-block weak-dependence structure for the standardized input vector $\cbarBm$ described in~\ref{blockstructure}
We establish Gaussian limits for the linear statistics $U_i$ and $\nu_m$ and for the nonlinear statistics $\lambda_k$.

\subsection{Linear Statistics $U_i$ and $\nu_m$}
Let $F\in\mathbb R^{D\times N}$ and $\widetilde W_m\in\mathbb R^D$ be fixed. Define
\begin{align}
    U_i &\defeq \frac{1}{\sqrt{D}} \sum_{r=1}^D F_{r,i} \cbarBm_r, \quad \text{for } i=1, \dots, N, \\
    \nu_m &\defeq \frac{1}{\sqrt{D}} \sum_{r=1}^D \widetilde{W}_{m,r} \cbarBm_r. \quad \text{for } m=1, \dots, M
\end{align}
For fixed $i$,
\begin{equation}
T_b^{(U)} \defeq \sum_{r \in \mathcal{B}_b} \frac{F_{r,i}}{\sqrt{D}} \cbarBm_r.
\end{equation}

By independence of the blocks $\CbarBm_{\mathcal{B}_b}$, the variables $\{T_b^{(U)}\}_b$ are independent with $\E T_b^{(U)}=0$ and $\sum_b\Var(T_b^{(U)})=\Var(U_i)\defeq\sigma_{U,i}^2$.

\subsubsection{Bound Property of Moments}\label{boundproperty}

Let assume that
\begin{assumption}[Bound assumption of fourth moment]\label{eq:A4}
here exists $K_4<\infty$ such that for all $b$ and all targets ($i$ for $U$, $m$ for $\nu$),
\begin{equation}
\E\!\big[(T_b^{(U)})^4\big] \;\le\; K_4\big(\Var(T_b^{(U)})\big)^2
\end{equation}
\end{assumption}
This holds, e.g., for Gaussian or sub-Gaussian blocks, or whenever all block-linear forms a uniform kurtosis bound.

By Cauchy-Schwarz,
\begin{equation}\label{eq:third-from-fourth}
\begin{aligned}
\E|T_b^{(U)}|^3
&\le\; \big(\E (T_b^{(U)})^2\big)^{1/2}\,\big(\E (T_b^{(U)})^4\big)^{1/2} \\
&\le\; \sqrt{K_4}\,\big(\Var(T_b^{(U)})\big)^{3/2}.
\end{aligned}
\end{equation}

Under the spectral bounds $c_\Sigma I\preceq\Sigma_b\preceq C_\Sigma I$,
\begin{equation}
    \Var(T_b^{(U)}) = \cfrac{1}{D}F_b\Sigma_bF_b^\top \;\le\;\cfrac{C_\Sigma}{D}||F_b||^2.
\end{equation}

Assume that
\begin{assumption}[Block bound assumption 2]\label{assumption:energy2}
For some $\beta_U>0$
\begin{equation}
\Var(T_b^{(U)})\le\;\cfrac{C_\Sigma}{D}||F_b||^2 \leq C_E (m/D)^{\beta_U}
\end{equation}
\end{assumption}
similarly as in Assumption~\ref{assumption:energy}.

As a result,
\begin{equation}\label{eq:sum-third-raw}
\sum_{b=1}^{n_b}\E|T_b^{(U)}|^3
\;\le\;\sqrt{K_4}\,n_b C_E^{3/2}\Big(\frac{m}{D}\Big)^{3\beta_U/2}.
\end{equation}

Since
\begin{equation}
    \frac{c_\Sigma}{D}\|F_{\cdot,i}\|^2 \ \le\ \sigma_{U,i}^2\ \le\ \frac{C_\Sigma}{D}\|F_{\cdot,i}\|^2
\end{equation}
from spectral bounds, using $n_b\asymp D/m$, we obtain
\begin{equation}
\sum_{b=1}^{n_b}\E\Big[\Big|\frac{T_b^{(U)}}{\sigma_{U,i}}\Big|^3\Big]
\;\le\; \sqrt{K_4}\\,
\Big(\frac{C_E D}{c_\Sigma\,\|F_{\cdot,i}\|^2}\Big)^{3/2}
\Big(\frac{m}{D}\Big)^{3\beta_U/2-1}.
\end{equation}

If we consider normalized $F$, $\|F_{\cdot,i}\|^2=D$, then 
\begin{equation}\label{eq:sum-third-final}
\sum_{b=1}^{n_b}\E\Big[\Big|\frac{T_b^{(U)}}{\sigma_{U,i}}\Big|^3\Big]
\le\; \frac{\sqrt{K_4}\,C_E^{3/2}}{c_\Sigma^{3/2}}\,
\Big(\frac{m}{D}\Big)^{3\beta_U/2-1}.
\end{equation}
\subsubsection{Gaussianity by Berry--Esseen Bound}
The Berry--Esseen theorem can be expressed by below.
\begin{theorem}[Berry--Esseen Bound]\label{berry}
The Berry--Esseen theorem for a sum of independent, non-identically distributed, zero-mean random variables $X_b$ states that if $\sum_b \Var(X_b) = 1$, then:
\begin{equation}
\sup_{x \in \R} | P(\sum_b X_b \le x) - \Phi(x) | \le C_{BE} \sum_b \mathbb{E}[|X_b|^3],
\end{equation}
where $\Phi(x)$ is the standard normal CDF and $C_{BE}$ is a universal constant.
\end{theorem}

Let $Z_b\defeq T_b^{(U)}/\sigma_{U,i}$ or $Z_b\defeq T_b^{(\nu)}/\sigma_{\nu,m}$, so $\sum_b\Var(Z_b)=1$. 
Berry--Esseen theorem implies followings.
\begin{theorem}
\label{thm:U_normality_BE}
There exists a constant $C_U$, independent of $D$ and $m$, such that for the Kolmogorov-Smirnov distance $d_{\mathrm{KS}}(\cdot, \cdot)$:
\begin{equation}
d_{\mathrm{KS}}\left( \frac{U_i}{\sigma_{U,i}}, \mathcal{N}(0,1) \right) \le C_U\Bigl(\frac{m}{D}\Bigr)^{3\beta_{U}/2-1}.
\end{equation}
An identical argument holds for $\nu_m$, yielding:
\begin{equation}
d_{\mathrm{KS}}\left( \frac{\nu_m}{\sigma_{\nu,m}}, \mathcal{N}(0,1) \right) \le C_\nu \Bigl(\frac{m}{D}\Bigr)^{3\beta_{\nu}/2-1}.
\end{equation}
where $\sigma_{\nu,m}^2 = \Var(\nu_m)$.
\end{theorem}
\subsection{Non-Linear Statistics $\lambda_k$}\label{sec:lambda-BE}
Recall, the non-linear linear statistic defined as
\begin{equation}\label{eq:lambda-def-sec}
  \lambda_k \;=\; \frac{1}{\sqrt N}\sum_{i=1}^{N} W_{k,i}\,f(U_i),
\end{equation}
with
\begin{equation}
    U_i \;=\; \frac{1}{\sqrt D}\sum_{r=1}^{D} F_{r,i}\,\cbarBm_r,
\end{equation}
Let $Z\sim\mathcal N(0,1)$ and set
\begin{equation}
a:=\E f(Z),\qquad b:=\E[Z\,f(Z)].
\end{equation}
By symmetry of function, we may assume $\E f(U_i)=0$ (otherwise replace $f$ by $f-\E f(U_i)$). 
\subsubsection{Gaussian Projections and Partial Gaussianity}
Define the Gaussian projection remainder
\begin{equation}
    r_i \;:=\; f(U_i) - a - b\,U_i
\end{equation}
The remainder covariance $\Cov(r_i,r_j)$ is
\begin{equation}\label{eq:ri-cov}
\Cov(r_i, r_j) < C_R \left(\frac{m}{D}\right)^\gamma.
\end{equation}
Let split $\lambda_k$ into linear main term and small remainder,
write
\begin{equation}
    \lambda_k \;=\; b\,L_k \;+\; R_k,
\end{equation}
where
\begin{equation}
    \begin{aligned}
L_k := \frac{1}{\sqrt N}\sum_{i=1}^N W_{k,i}\,U_i,\\
R_k := \frac{1}{\sqrt N}\sum_{i=1}^N W_{k,i}\,r_i .
    \end{aligned}
\end{equation}

For $L_k$, $L_k$ itself is a linear form in $(\overline C_r)$
\begin{equation}
    L_k \;=\; \sum_{r=1}^{D} \alpha_r\,\cbarBm_r, \quad
\alpha_r \;:=\; \frac{1}{\sqrt{ND}}\sum_{i=1}^{N} W_{k,i}\,F_{r,i}.
\end{equation}
$L_k$ also calculated by summation of block groups  $\{\mathcal B_b\}_{b=1}^{n_b}$,
\begin{equation}
Y_b \;:=\; \sum_{r\in\mathcal B_b} \alpha_r\,\cbarBm_r,
\quad
L_k \;=\; \sum_{b=1}^{n_b} Y_b,\quad \E Y_b=0.
\end{equation}
Independence in $\{Y_b\}$, we can apply Berry--Esseen theorem to $\sum_b (Y_b/\sigma_{L_k})$
\begin{equation}
    d_{\mathrm{KS}}\!\left(\frac{L_k}{\sigma_{L_k}},\,\mathcal N(0,1)\right)
\;\le\; C_{BE}\sum_{b=1}^{n_b}\E\!\left[\left|\frac{Y_b}{\sigma_{L_k}}\right|^3\right].
\end{equation}
where $\sigma_{L_k}:=\sqrt{\Var(L_k)}$.
Along similar procedure with Appendix~\ref{boundproperty}, we can get bound property 
\begin{equation}
\sum_{b=1}^{n_b}\E|Y_b|^3
\;\le\; C\,\Big(\frac{m}{D}\Big)^{3\beta_L/2-1}
\end{equation}

Therefore
\begin{equation}\label{eq:BE-Lk}
d_{\mathrm{KS}}\!\left(\frac{L_k}{\sigma_{L_k}},\,\mathcal N(0,1)\right)
\;\le\; C_L\, \Bigl(\frac{m}{D}\Bigr)^{3\beta_{L}/2-1}.
\end{equation}
\subsubsection{Gaussianity by Berry--Esseen in Splits}
From \eqref{eq:ri-cov},
\begin{equation}
\Var(R_k)
=\frac{1}{N}\sum_{i,j}W_{k,i}W_{k,j}\Cov(r_i,r_j)
\;\le\; C_2\,(m/D)^{\gamma}.
\end{equation}
Using the following lemma,
\begin{lemma}
For any real r.v.'s $X,Y$ and any $t>0$,
\begin{equation}
    d_{\mathrm{KS}}(X+Y,\;\mathcal N)\ \le\ d_{\mathrm{KS}}(X,\mathcal N)\ +\ P(|Y|>t)\ +\ c\,t,
\end{equation}
with $c=\sqrt{2/\pi}$
\end{lemma}
where can directly derived from event set relationship.
In particular, by Chebyshev $P(|Y|>t)\le \Var(Y)/t^2$,
\begin{equation}
d_{\mathrm{KS}}(X+Y,\;\mathcal N)
\ \le\ d_{\mathrm{KS}}(X,\mathcal N)\ +\frac{\Var(Y)}{t^2}+c\,t.
\end{equation}
Choosing $t=(\Var(Y)/c)^{1/3}$ gives the clean bound
\begin{equation}\label{eq:smooth-cuberoot}
d_{\mathrm{KS}}(X+Y,\;\mathcal N)\ \le\ d_{\mathrm{KS}}(X,\mathcal N)\ +\ C_3\;\Var(Y)^{1/3}.
\end{equation}

Kolmogorov distance is invariant under nonzero affine transforms:
for $a\neq0$, $b\in\R$,
\begin{equation}
d_{\mathrm{KS}}(aX+b,\,aY+b)=d_{\mathrm{KS}}(X,Y).
\end{equation}
Let $\sigma_{\lambda_k}^2=\Var(\lambda_k)>0$.
Apply \eqref{eq:smooth-cuberoot} with $X=bL_k$ and $Y=R_k$ against the Gaussian
$\mathcal N_b:=\mathcal N(0,b^2\sigma_{L_k}^2)$,
\begin{equation}
\begin{aligned}
    d_{\mathrm{KS}}(bL_k+R_k,\mathcal N_b)
\ \le\ &d_{\mathrm{KS}}(bL_k,\mathcal N_b)\\&+C_3\,\Var(R_k)^{1/3}.
\end{aligned}
\end{equation}
The first term on right-hand side satisfies
\begin{equation}
d_{\mathrm{KS}}(bL_k,\mathcal N_b)=d_{\mathrm{KS}}\!\Big(\frac{L_k}{\sigma_{L_k}},\mathcal N\Big)
\ \le\ C_L\Bigl(\frac{m}{D}\Bigr)^{3\beta_L/2-1},
\end{equation}
results to
\begin{equation}
d_{\mathrm{KS}}(\lambda_k,\mathcal N_b))
\ \le\ C_L\Bigl(\frac{m}{D}\Bigr)^{3\beta_L/2-1}+C_3\,\Var(R_k)^{1/3}.
\end{equation}

Compare $\mathcal N(0,b^2\sigma_{L_k}^2)$ with $\mathcal N(0,\sigma_{\lambda_k}^2)$, which distance between these two Gaussians is $\mathcal O(\sqrt{\Var(R_k)})$, and can be absorbed into the other dominate term.

Collecting the bounds, $\Var(R_k)\le C_2 (m/D)^{\gamma}$,
\begin{equation}\label{eq:lambda-rate-final}
d_{\mathrm{KS}}\!\Big(\frac{\lambda_k}{\sigma_{\lambda_k}},\mathcal N(0,1)\Big)
\ \le\
C_{\lambda,1}\Bigl(\frac{m}{D}\Bigr)^{\beta_L/2}
\ +\
C_{\lambda,2}\Bigl(\frac{m}{D}\Bigr)^{\gamma/3}
\end{equation}

\section{Scaling law in the Gaussian limit}\label{appendix:GaussianLimit}
In our derivations of the Function Correlation Decomposition and the Gaussianity results, we used block moment bounds such as Assumptions~\ref{assumption:energy} and~\ref{assumption:energy2}.
When the feature matrix $F$ is generated from a Gaussian model and then column-wise normalized (i.e., $\|F_{\cdot,i}\|^2=D$ for all $i$), we can identify the exponent $\beta$ in the ideal limit. 
For simplicity we focus on the statistics $T_b^{(U)}$ associated with $F$; the same argument applies to $T_b^{(\nu)}$ in the derivation of $\nu$ and to $Y_b$ in the derivation of $\lambda$.

Fix a block partition $\{\mathcal B_b\}_{b=1}^{n_b}$ of $\{1,\dots,D\}$ with $|\mathcal B_b|=m$ and $n_b=D/m$ (assume $m\mid D$ for notational convenience).
Assume each column of $F$ is drawn as
\begin{equation}
F_{\cdot,i}\;=\;\sqrt{D}\,\frac{G}{\|G\|},\qquad G\sim\mathcal N(0,I_D),
\end{equation}
so that $\|F_{\cdot,i}\|^2=D$ almost surely.

By exact partitioning,
\begin{equation}\label{eq:block-average}
\frac{1}{n_b}\sum_{b=1}^{n_b}\|F_{\mathcal B_b,i}\|^2
\;=\;\frac{1}{n_b}\,\|F_{\cdot,i}\|^2
\;=\; m.
\end{equation}
Consider $\chi^2$ distribution, the tail condition arises.
In detail, there exists $c>0$ such that for all $t>0$,
\begin{equation}\label{eq:tail-block-energy}
\Pr\!\left(\left|\frac{\|F_{\mathcal B_b,i}\|^2}{m}-1\right|\ge t\right)
\;\le\; 2e^{-c\,m\min\{t^2,t\}}\;+\;2e^{-c\,D\min\{t^2,t\}}.
\end{equation}

In other words, for any block $\mathcal B_b$,
\begin{equation}\label{eq:chi2-ratio}
\frac{\|F_{\mathcal B_b,i}\|^2}{m}
\;=\;\frac{D}{m}\cdot\frac{\chi^2_m}{\chi^2_D}
\;=\; 1 \pm O_{p}\!\Big(m^{-1/2}+D^{-1/2}\Big)
\end{equation}
where $O_p(r)$ denotes a term bounded in probabililty at rate $r$.
In particular, if $m\to\infty$ and $D\to\infty$ with $\rho:=m/D\in(0,1]$ fixed, then
\begin{equation}\label{eq:block-energy-whp}
\|F_{\mathcal B_b,i}\|^2 \;=\; m\,(1\pm \varepsilon_{m,D}).
\end{equation}

Combining \eqref{eq:chi2-ratio} with the spectral bounds $c_\Sigma I\preceq \Sigma_b\preceq C_\Sigma I$, for each block
\begin{equation}\label{eq:VarTbU-upper}
\Var\!\big(T_b^{(U)}\big)
\;\le\;\frac{C_\Sigma}{D}\,\|F_{\mathcal B_b,i}\|^2
\;=\; C_\Sigma\,\frac{m}{D}\,\big(1\pm \varepsilon_{m,D}\big),
\end{equation}
where $\varepsilon_{m,D}$ is as in \eqref{eq:block-energy-whp}. In the joint limit $m\to\infty$ and $D\to\infty$, fluctuation $\varepsilon_{m,D}$ vanishes with high probability.
Therefore the block-moment scaling exponent in Assumption~\ref{assumption:energy}, Assumption~\ref{assumption:energy2} may be taken as $\beta\to1$.

Consequently, all Berry--Esseen bounds that depend on $\beta$ specialize to the rate
\begin{equation}
d_{\mathrm{KS}}\!\left(\frac{U_i}{\sigma_{U,i}},\mathcal N(0,1)\right)
\ \le\ C'\left(\frac{m}{D}\right)^{1/2}.
\end{equation}

\section{Additional Information in Pseudo-Real Dataset Experimental Setting}\label{section:discussion:mnistresults}
In this pseudo-real dataset investigation, the dimension of teacher model input $C$ was set to $D=500$, and the dimension of student model input $X$ was set to $N=28^2$, MNIST dimension.
The encoder model consisted of a hidden layer with 256 units and used the ReLU activation function. The teacher and student models followed the same architecture as in the Gaussian mixture setting, with both having hidden layer dimensions set to $K=M=2$. Both the teacher and student models employed the same function, $\text{hardtanh}(x) = \max(-1, \min(1, x))$. 
The nonlinear function $f(x)$ used to generate the student input was also defined as $f(\cdot)=\text{hardtanh}(\cdot)$.
During training, we used 10,000 samples of teacher model input and employed a full-batch Adam optimizer for optimization with $10^{-4}$ learning rate. For the SGD and ODE dynamics simulations, we used 100,000 samples, generated using the encoder model applied to the MNIST dataset. Training was conducted for 30 epochs, and at each epoch, we recorded the corresponding teacher model and teacher model input pairs. To analyze the effect of correlation strength, we repeated the training process for different values of $\lambda_{corr}$, varying it over the ranges $\lambda_{corr}=[1.1,1.2,1.3,1.4,1.5,1.6,1.7,1.8,1.9]$ and $\lambda_{corr}=[1,2,3,4,5,6,7,8,9,10]$.

\newpage
\begin{widetext}
\section{Additional Dynamics Results}
\FloatBarrier
\begin{figure}
\centering
\begin{tabular}{cc}
\includegraphics[width=0.5\textwidth ]{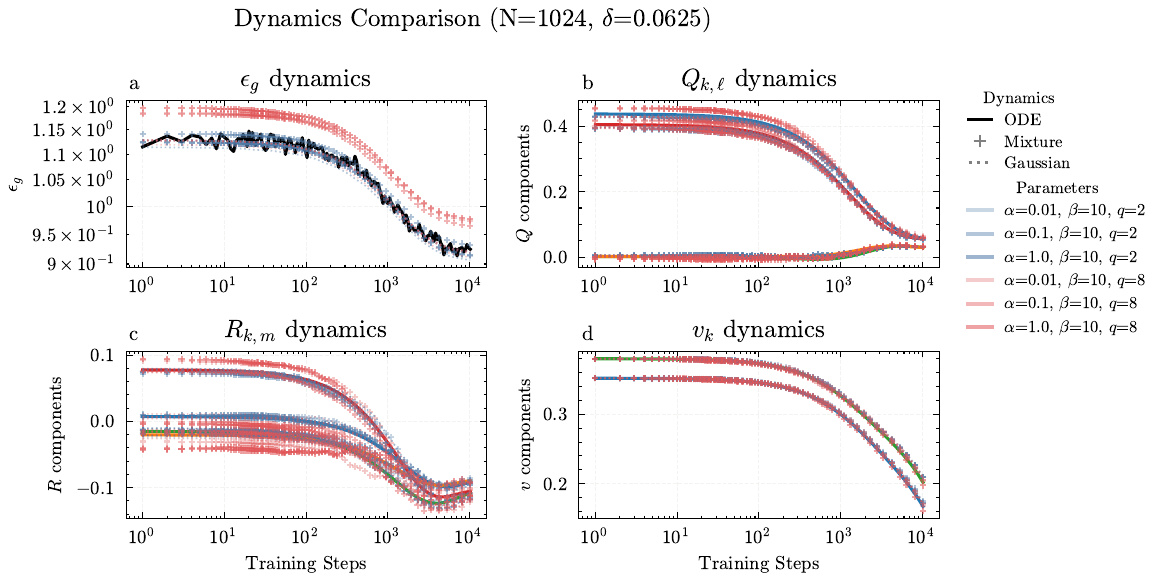}& \includegraphics[width=0.5\textwidth ]{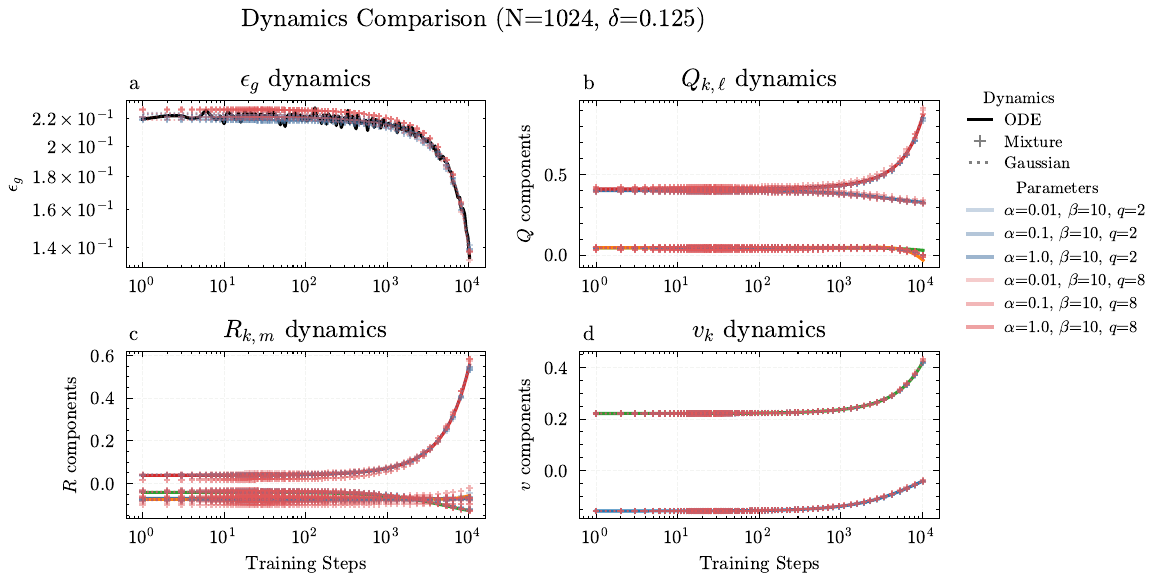}\\ \includegraphics[width=0.5\textwidth ]{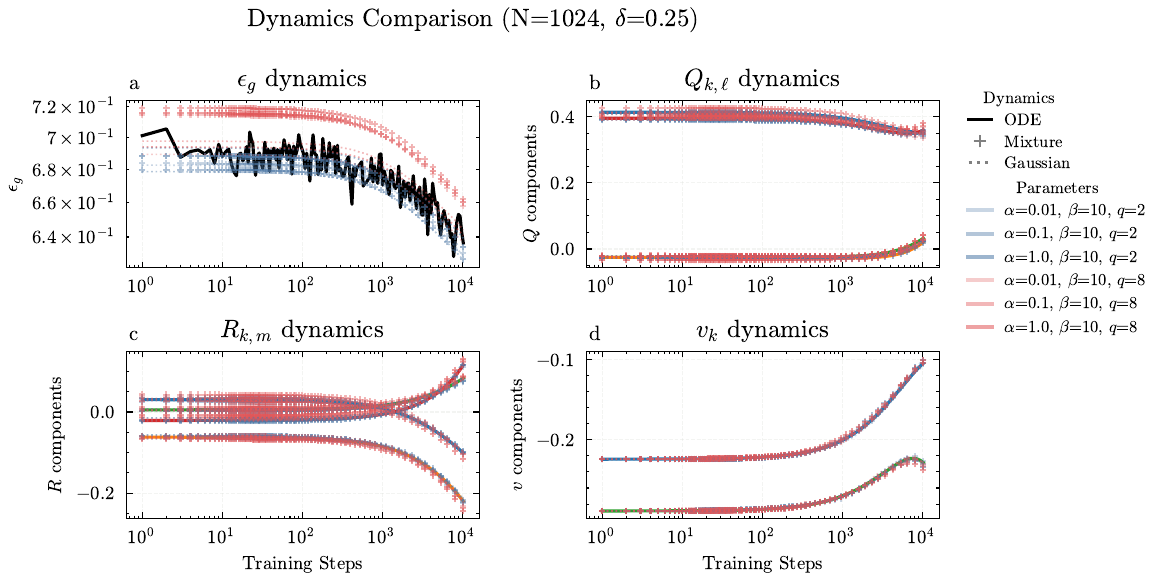}& \includegraphics[width=0.5\textwidth ]{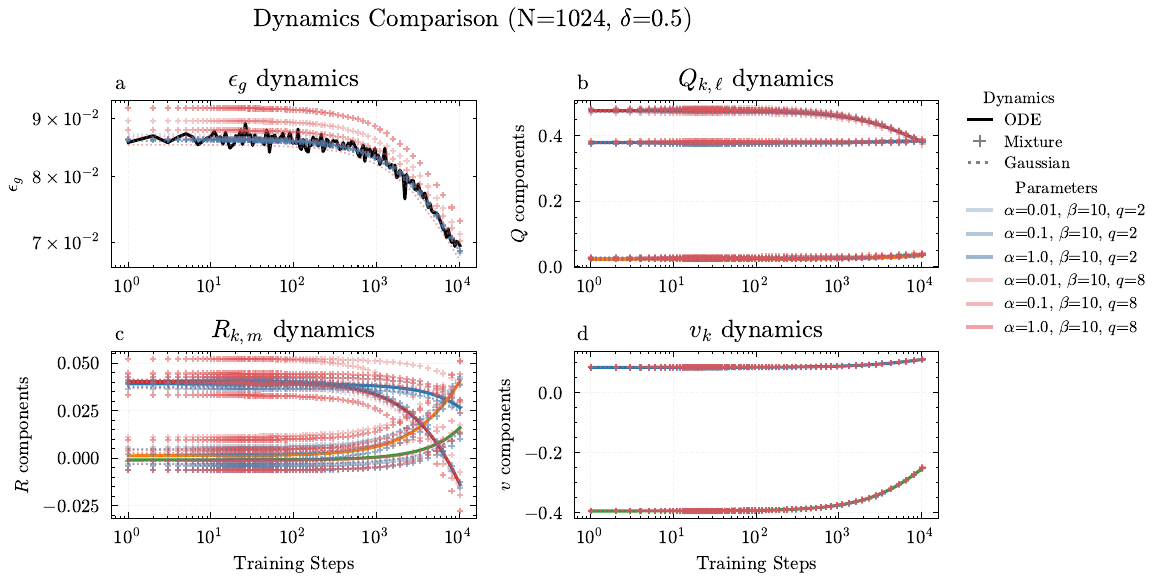}\\
\end{tabular}
\caption{Dynamics Comparison in $N=1024$}
\end{figure}

\begin{figure}
\centering
\begin{tabular}{cc}
\includegraphics[width=0.5\textwidth ]{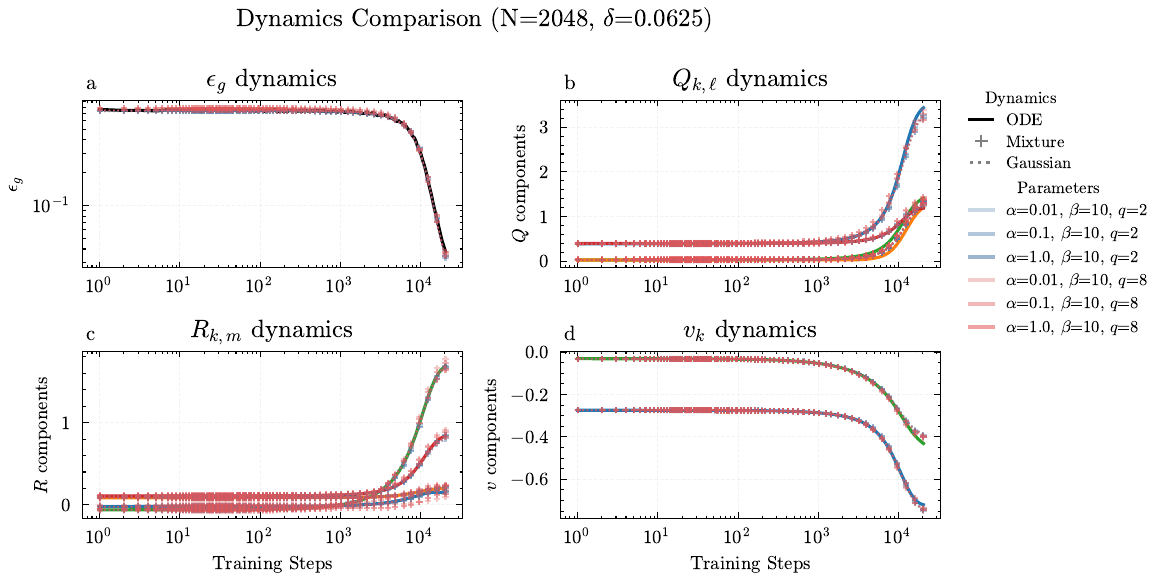}& \includegraphics[width=0.5\textwidth ]{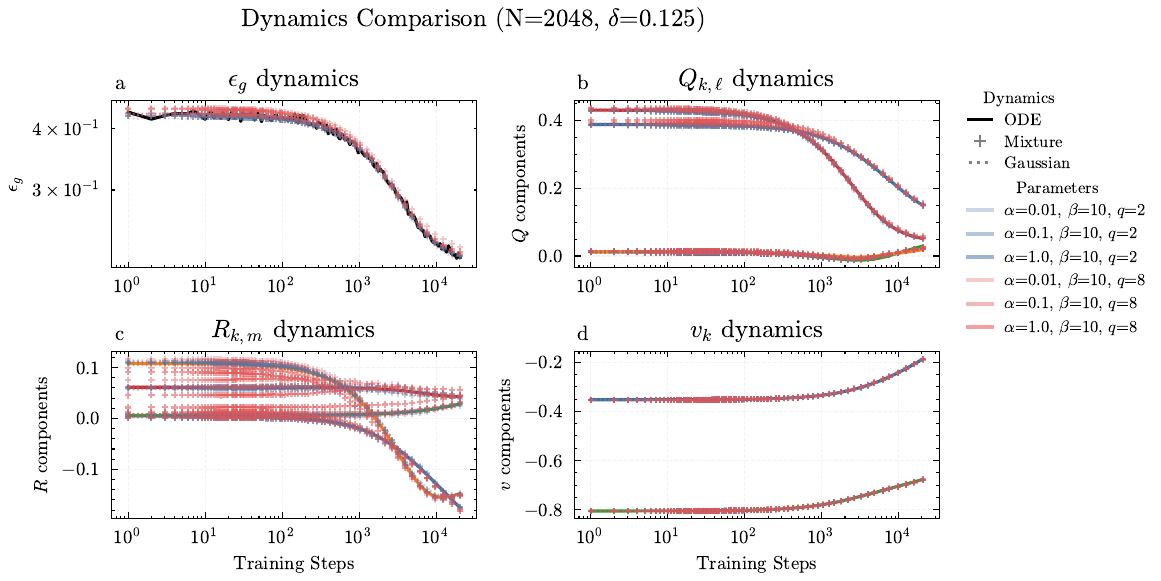}\\ \includegraphics[width=0.5\textwidth ]{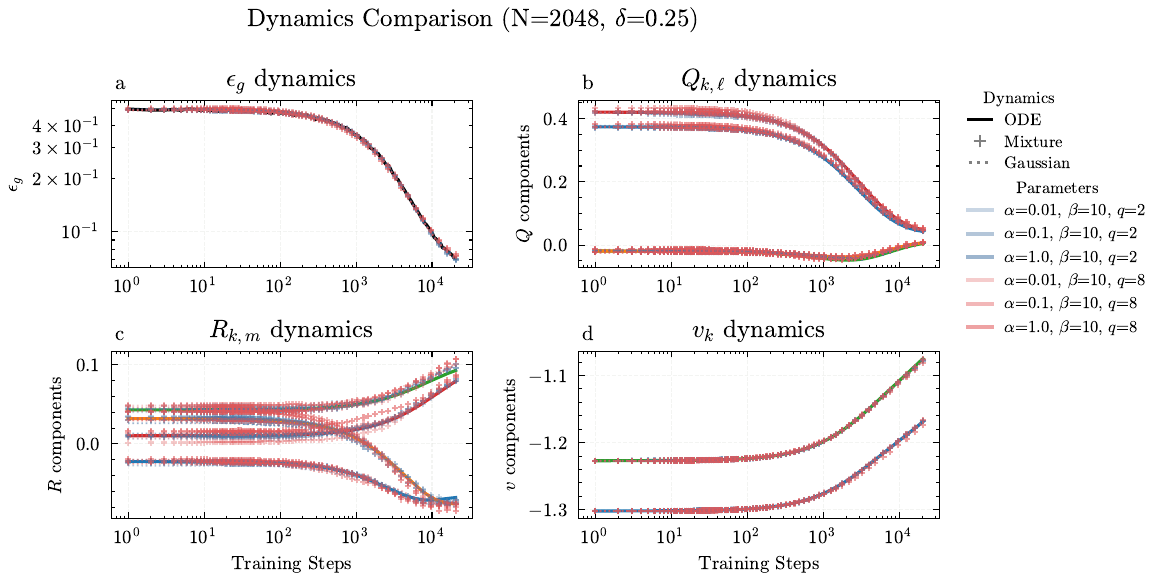}& \includegraphics[width=0.5\textwidth ]{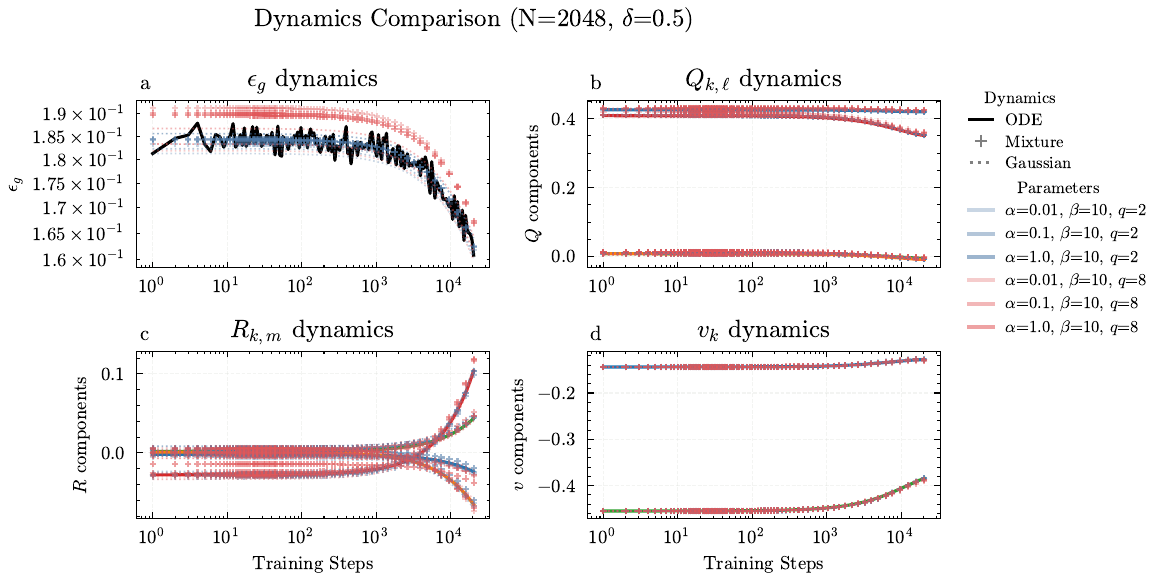}\\
\end{tabular}
\caption{Dynamics Comparison in $N=2048$}
\end{figure}

\begin{figure}
\centering
\begin{tabular}{cc}
\includegraphics[width=0.5\textwidth ]{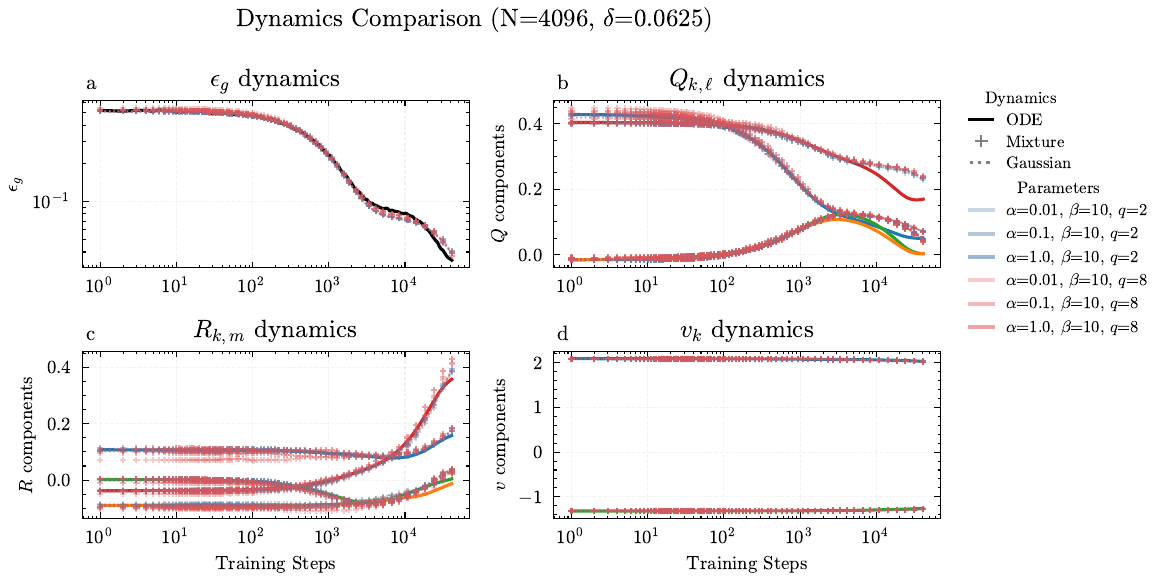}& \includegraphics[width=0.5\textwidth ]{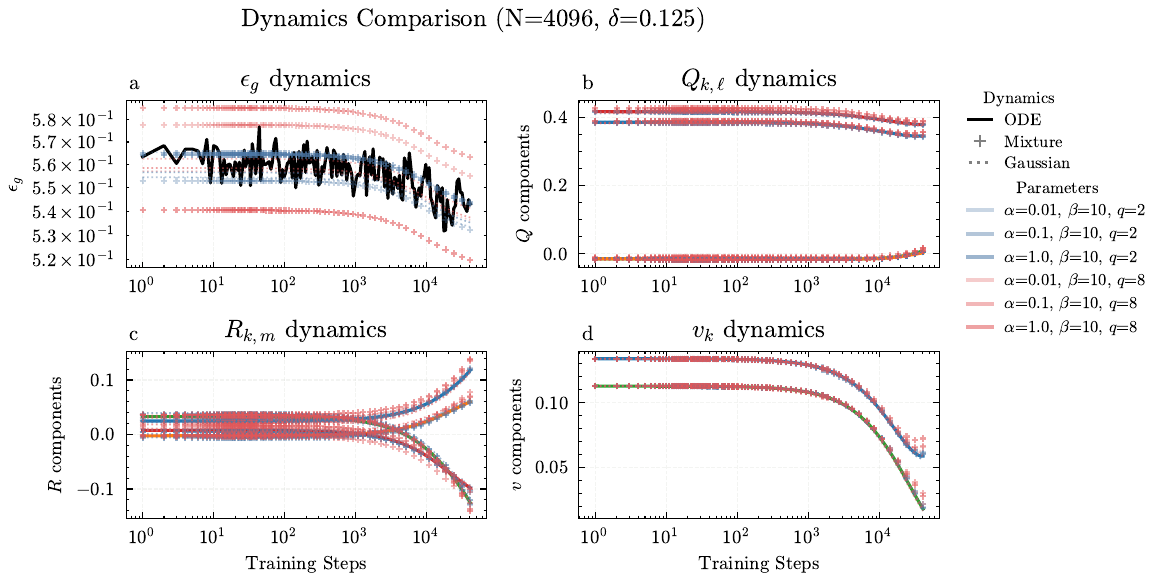}\\ \includegraphics[width=0.5\textwidth ]{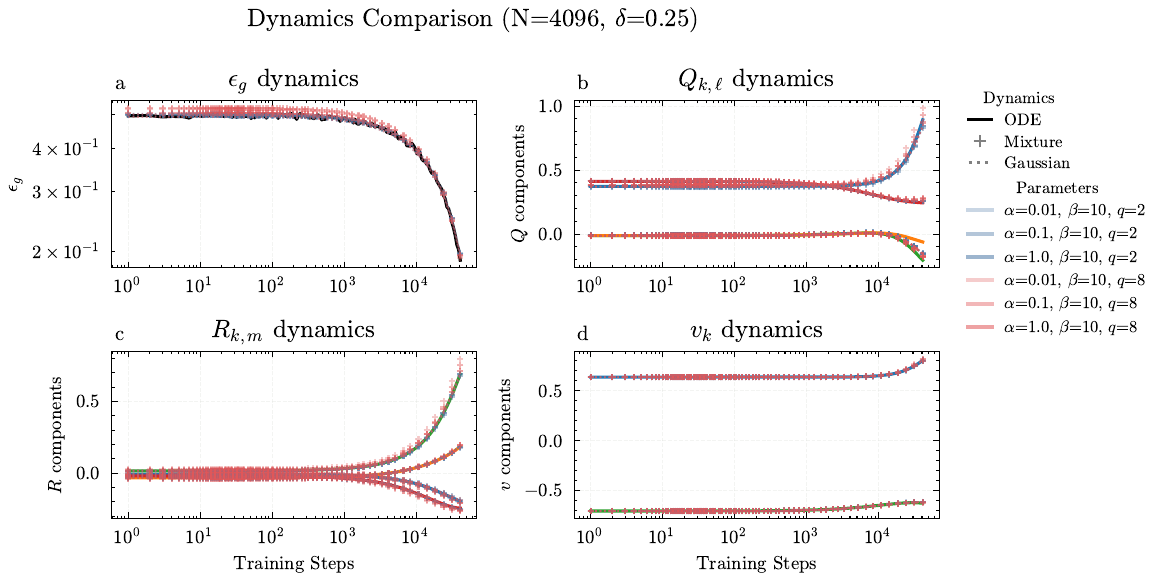}& \includegraphics[width=0.5\textwidth ]{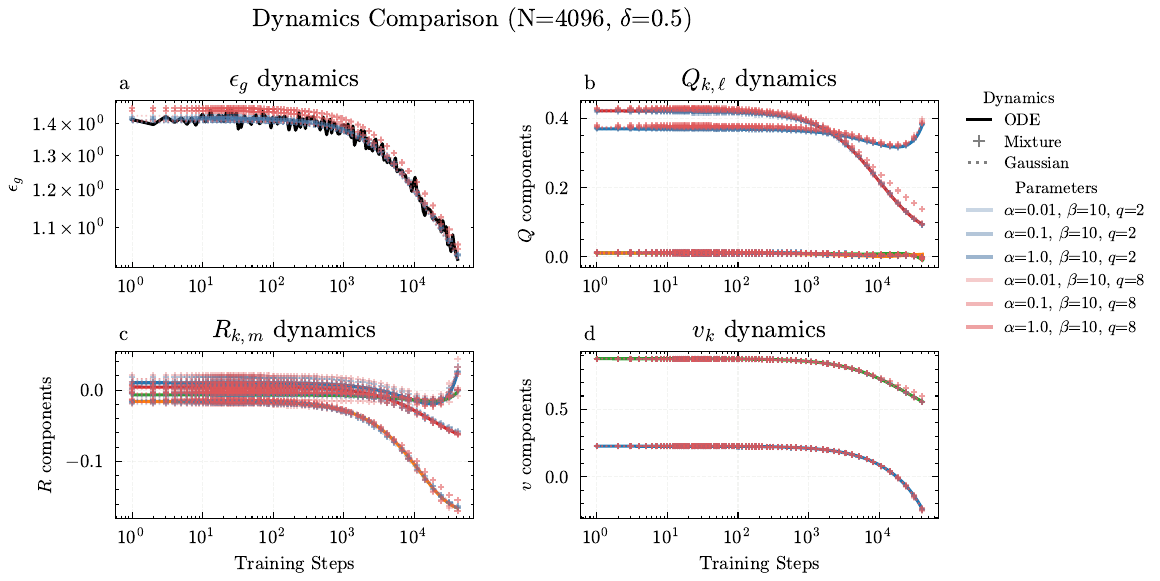}\\
\end{tabular}
\caption{Dynamics Comparison in $N=4096$}
\end{figure}
\end{widetext}
\FloatBarrier
\newpage

\bibliographystyle{apsrev4-2}
\bibliography{PRE-GU-BJ}
\end{document}